\documentclass{article}

\usepackage[utf8]{inputenc} 
\usepackage[T1]{fontenc}    
\usepackage{microtype,inconsolata}

\usepackage{algorithmic,algorithm}

\usepackage[utf8]{inputenc}

\usepackage{graphicx} \graphicspath{{figures/}}
\usepackage{booktabs,tabularx,colortbl,multirow,multicol,array,makecell,tabularray}
\usepackage{subcaption}
\usepackage{caption}

\usepackage[switch]{lineno}
\usepackage{threeparttable}

\usepackage{amsmath,amssymb,mathtools,amsthm,nicefrac}

\usepackage{acronym}
\usepackage{enumitem}
\usepackage{balance}
\usepackage{xspace}
\usepackage{setspace}

\usepackage[hidelinks,pagebackref,breaklinks,colorlinks,citecolor=gblue,linkcolor=blue]{hyperref} 
\usepackage[capitalise,noabbrev,nameinlink]{cleveref}

\theoremstyle{plain}

\theoremstyle{definition}

\theoremstyle{remark}

\usepackage{overpic,wrapfig}
\usepackage[misc]{ifsym}
\usepackage{pifont}
\usepackage{natbib}

\makeatletter
\DeclareRobustCommand\onedot{\futurelet\@let@token\@onedot}
\def\@onedot{\ifx\@let@token.\else.\null\fi\xspace}

\makeatother

\crefname{algorithm}{Alg.}{Algs.}
\Crefname{algocf}{Algorithm}{Algorithms}
\crefname{section}{Sec.}{Secs.}
\Crefname{section}{Section}{Sections}
\crefname{table}{Tab.}{Tabs.}
\Crefname{table}{Tab.}{Tabs.}
\crefname{figure}{Fig.}{Figs.}
\Crefname{figure}{Fig.}{Figs.}
\crefname{equation}{Eq.}{Eqs.}
\Crefname{equation}{Equation}{Equations}
\crefname{appendix}{Appx.}{Appxs.}
\Crefname{appendix}{Appendix}{Appendices}

\acrodef{INSIGHT}[INSIGHT]{interpretable neuro-symbolic framework for reinforcement Learning with textual explanation}

\usepackage[accepted]{icml2024}
\usepackage{tabularx} 
\usepackage{longtable}
\definecolor{lightgray}{rgb}{0.9, 0.9, 0.9}  
\icmltitlerunning{End-to-End Neuro-Symbolic Reinforcement Learning with Textual Explanations}

\begin{document}

\twocolumn[
\icmltitle{End-to-End Neuro-Symbolic Reinforcement Learning with Textual Explanations}



\icmlsetsymbol{equal}{*}

\begin{icmlauthorlist}
\icmlauthor{Lirui Luo}{sch,comp}
\icmlauthor{Guoxi Zhang}{comp}
\icmlauthor{Hongming Xu}{comp}
\icmlauthor{Yaodong Yang}{sch,comp}
\icmlauthor{Cong Fang}{sch}
\icmlauthor{Qing Li}{comp}
\end{icmlauthorlist}

\icmlaffiliation{sch}{
School of Intelligence Science and Technology, Peking University}
\icmlaffiliation{comp}{State Key Laboratory of General Artificial Intelligence, BIGAI}

\icmlcorrespondingauthor{Qing Li}{dylan.liqing@gmail.com}
\icmlcorrespondingauthor{Cong Fang}{fangcong@pku.edu.cn}

\icmlkeywords{Machine Learning, ICML}

\vskip 0.3in
]

\printAffiliationsAndNotice{Project page: \href{https://ins-rl.github.io}{ins-rl.github.io}.} 


\begin{abstract}
Neuro-symbolic reinforcement learning (NS-RL) has emerged as a promising paradigm for explainable decision-making, characterized by the interpretability of symbolic policies.
NS-RL entails structured state representations for tasks with visual observations, but previous methods cannot refine the structured states with rewards due to a lack of efficiency.
Accessibility also remains an issue, as extensive domain knowledge is required to interpret symbolic policies.
In this paper, we present a neuro-symbolic framework for jointly learning structured states and symbolic policies, whose key idea is to distill the vision foundation model into an efficient perception module and refine it during policy learning.
Moreover, we design a pipeline to prompt GPT-4 to generate textual explanations for the learned policies and decisions, significantly reducing users' cognitive load to understand the symbolic policies.
We verify the efficacy of our approach on nine Atari tasks and present GPT-generated explanations for policies and decisions.
\end{abstract}

\section{Introduction}\label{sec:intro}

Recent years have witnessed remarkable progress of deep reinforcement learning (RL)~\citep{agarwal2021deep,wurman2022outracing,degrave2022magnetic}.
However, deep RL still faces limitations in sensitive domains due to its opaque nature~\citep{milani2022survey}. For example, the opacity hinders diagnosing or rectifying policies that fails to generalize, which may be caused by exploiting peripheral information~\citep{delfosse2024interpretable}.
Neuro-symbolic reinforcement learning (NS-RL) is promising for overcoming this limitation~\citep{verma2018programmatically,coppens2019distilling}.
It uses structured states and parameterizes policies with concise expressions, thereby guaranteeing interpretability via clear semantics of states and policies.

For tasks with visual observations, the structured states need to be learned from pixels, which involves identifying objects from images.
\citet{zheng2022symbolic} proposed to use the Spatially Parallel Attention and Component Extraction (SPACE) model~\citep{Lin2020SPACE} for this purpose, but the state representations are fixed during policy learning due to the computational overhead of SPACE.
That being said, the structured states are not refined with reward signals, leading to significant performance degradation.

Meanwhile, symbolic policies can be intricate for general users, albeit being intrinsically interpretable.
For instance, to interpret logical policies~\citep{delfosse2023interpretable} one has to be familiar with first-order logic, and in the case of programmatic policies~\citep{qiu2022programmatic} one has to learn the corresponding grammars.
Such a lack of accessibility can be an obstacle for agents to gain the trust of the general public.
Nevertheless, there is a lack effort in the literature of NS-RL to explain policies for non-expert users.

\begin{figure}[t]
  \centering
  \includegraphics[width=\linewidth]{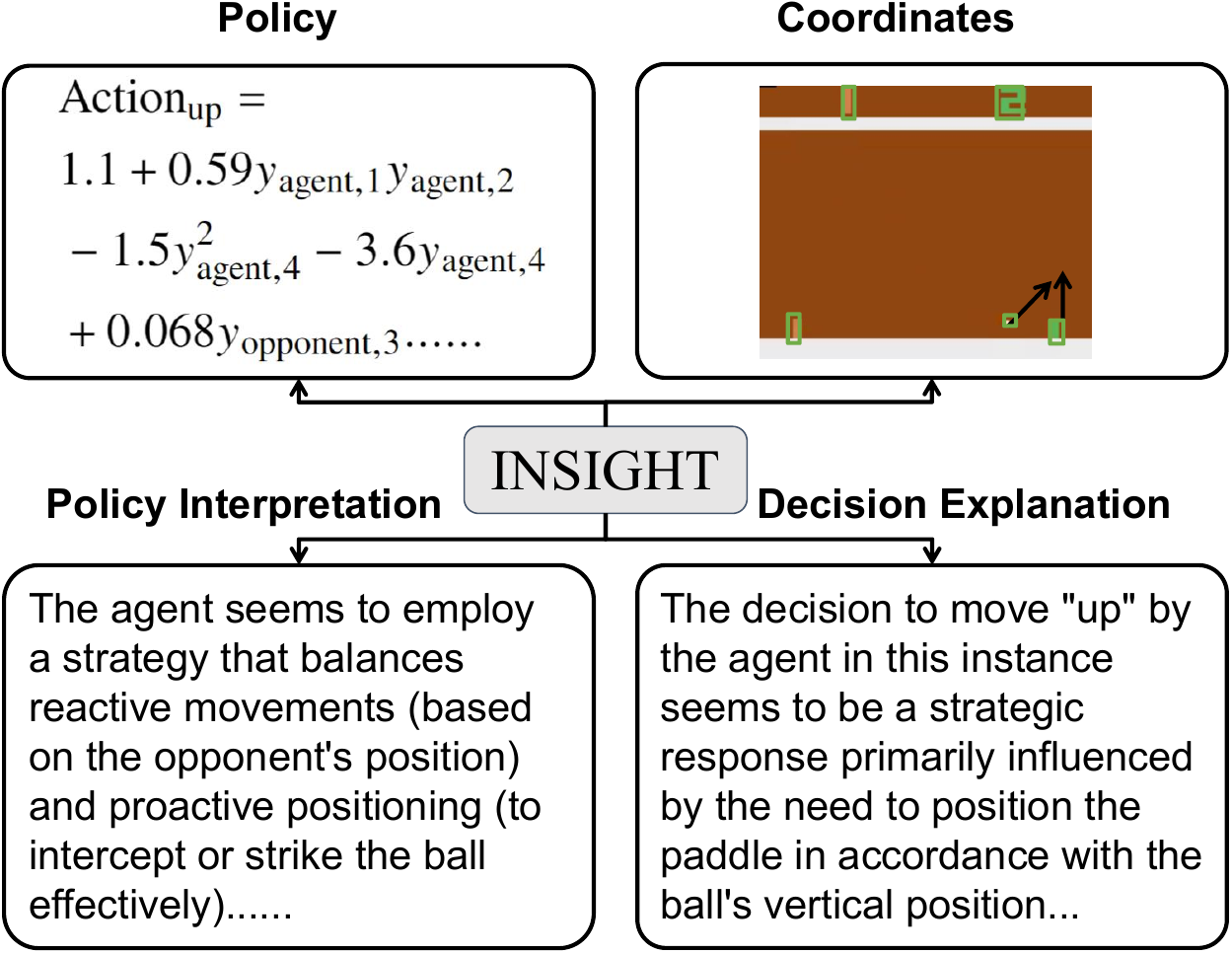}
    \caption{For tasks with visual input, \ac{INSIGHT} can simultaneously learn the coordinates of objects in observations and coordinate-based symbolic policies simultaneously, and it can interpret learned policies and specific decisions in natural language. $\text{y}_\text{{agent,i}}$ represents the vertical coordinate of the agent in the $i^{th}$ frame. Both policy interpretation and decision explanation are produced by entering the policy and a predefined prompt template into the LLM.}
\label{teaser}
\vskip -0.1in
\end{figure}

To address these issues, we present \acs{INSIGHT}, an interpretable neuro-symbolic framework for visual reinforcement learning.
As illustrated in \cref{teaser}, INSIGHT can learn the object coordinates and symbolic policies simultaneously and explain policies and specific decisions in natural language.
The key idea of INSIGHT is to overcomes the efficiency drawback of previous methods by distilling vision foundation models into a scalable perception module.
Accompanied by the equation learner (EQL)~\citep{sahoo2018learning} for representing policies, both the perception module and symbolic policies can be learned from rewards in an end-to-end fashion.
Moreover, we develop a pipeline for explaining policies and decisions in natural language via GPT-4, which reduces the cognitive load on users to understand symbolic policies learned by INSIGHT.
Through a step called concept grounding, LLMs are instructed to associate quantities in symbolic policies with their semantics for the task of interest.
Then, as illustrated in \cref{teaser}, LLMs are prompted for explaining the decision-making patterns of a symbolic policy and specific decisions made by the policy.

We verify the efficacy of INSIGHT and its model design with extensive experiments on nine Atari games. INSIGHT outperforms all existing approaches for NS-RL (CGP~\citep{wilson2018evolving}, Diffses~\citep{yuan2022pre}, DSP~\citep{landajuela2021discovering}, and NUDGE~\citep{delfosse2023interpretable}).
We show that its empirical performance can be attributed to the refinement of structured states with reward information and our approach for learning symbolic policies.
We also present examples of policy interpretations and decision explanations.

In summary, our contributions are three-fold.
\begin{enumerate}
    \item We propose an NS-RL framework that can refine structured states with both visual and reward information.
    \item We develop a pipeline to prompt GPT-4 to explain the learned policies and decisions in natural language.
    \item We demonstrate the efficacy of our framework on nine Atari games and showcase textual explanations.
\end{enumerate}

\section{Related work}
\label{gen_inst}

Approaches for explainable RL can be categorized into post hoc and intrinsic methods.
The former generates explanations using predefined templates~\citep{ehsan2018rationalization,wang2019verbal,hayes2017improving} or saliency maps~\citep{greydanus2018visualizing}, which can be subjective and unreliable~\citep{dazeley2023explainable}.
The latter relies on interpretable policy classes, such decision trees~\citep{topin2021iterative,zhang2021off}, logical expressions~\citep{dazeley2023explainable,delfosse2023interpretable,delfosse2024interpretable}, and mathematical expressions~\citep{zheng2022symbolic,landajuela2021discovering}, for better transparency.
Meanwhile, we argue that intrinsic transparency does not guarantee accessibility.
While logic-based approaches are better for characterizing relations, they can be unfriendly to users due to their prerequisite knowledge.
Recently, there is a growing interest in using LLMs to explain machine learning models \citep{kroeger2023large,tennenholtz2023demystifying,singh2023explaining,bills2023language,zhang2023explaining}.
INSIGHT is the first NS-RL method that improves accessibility by using LLMs to generate explanations automatically. 

While neuro-symbolic approaches \citep{manhaeve2018deepproblog,li2020ngs,li2024nsr} have better interpretability, prior NS-RL approaches~\citep{verma2018programmatically,coppens2019distilling,verma2019imitation,landajuela2021discovering} focus on tasks with low-dimensional observations that have clear semantics.
For tasks with visual input, they either rely on ground truth~\citep{delfosse2023interpretable} or human-defined primitives~\citep{wilson2018evolving,lyu2019sdrl} for state representations.
As an exception, \citet{zheng2022symbolic} proposed to extract structured states from pre-trained SPACE models, yet the representations are not refined with reward signals and lead to significant performance degradation.
INSIGHT is the first NS-RL approach that learns the structured states from both visual and reward signals.

\section{INSIGHT}\label{mehtod}
\begin{figure*}[ht!]
  \centering
  \includegraphics[width=0.9\textwidth]{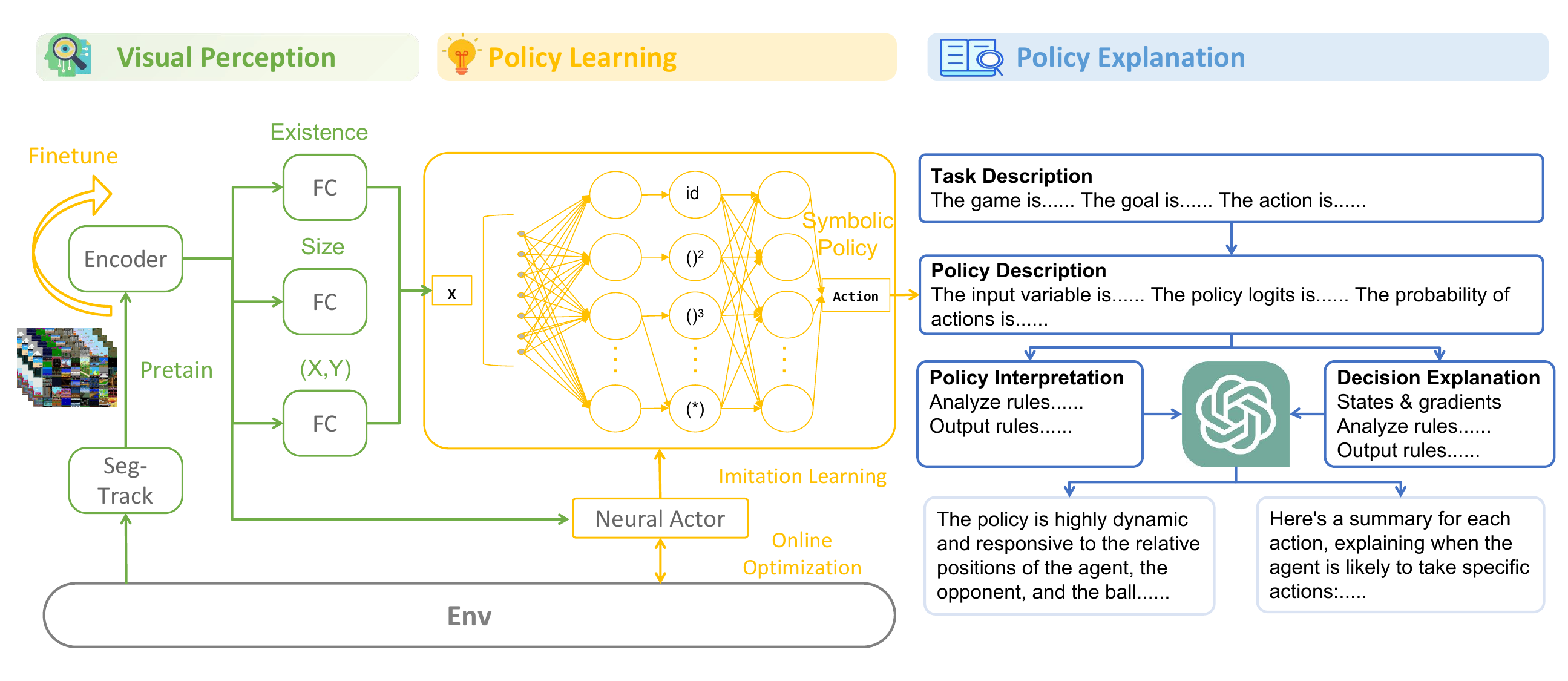}
    \caption{INSIGHT consists of three components: a perception module, a policy learning module, and a policy explanation module. The perception module learns to predict object coordinates using a frame-symbol dataset distilled from vision foundation models. The policy learning module is responsible for learning coordinate-based symbolic policies. In particular, to address with the limited expressiveness of object coordinates, it uses a neural actor to interact with the environment. The policy explanation module can generate policy interpretations and decision explanations using task description, policy description, and values of object coordinates.}\label{frame}
\vskip -0.1in
\end{figure*}

This section presents the proposed INSIGHT framework.
As illustrated in~\cref{frame}, it consists of three components: a visual perception module (\cref{subsec:state_learning}), a policy learning module (\cref{subsec:end_to_end}), and a policy explanation module (\cref{enhanced-interpretability}).

\subsection{Visual Perception}\label{subsec:state_learning}
\textbf{Frame-Symbol Dataset\quad}
The perception module needs to extract information about objects from input images.
Given the notorious sample efficiency of online RL, approaches that use image reconstruction objectives~\citep{zheng2022symbolic,yoon2023investigation} are prohibitively expansive.
We claim that, for NS-RL, compared to reproducing every detail in visual observations, it suffices to only recognize the positions of objects.
To this end, we consider harvesting the segmentation and tracking ability of vision foundation models.
For each task, we first roll out about 10,000 frames using pre-trained neural agents and then extract the object bounding boxes using the FastSAM segmentation model~\citep{zhao2023fast} and the DeAot model~\citep{yang2022decoupling}.
In specific, segmentation is repeated every ten frames to capture new objects, and the DeAot model is also capable of capturing unseen objects during tracking. 
Using the bounding boxes, we compute the coordinates of objects (i.e., their centers) and their width and height, normalize these quantities to $[0,1]$, and pair them with the corresponding images to form a frame-symbol dataset $\mathcal{D}_\text{symbol}$. 
Details of the dataset generation are provided in \cref{Enhanced Description of Frame-Symbol Dataset Generation}. 
By fitting to this dataset, the perception module can learn to extract the structured states before policy learning, which can improve the sample efficiency of INSIGHT.

\textbf{Learning Object Coordinates\quad}
We now present the details of learning from $\mathcal{D}_\text{symbol}$.
$\mathcal{D}_\text{symbol}$ contains information of the existence, the shape, and the coordinates of objects.
To make full use of them, INSIGHT employs a multi-task formulation for the perception module $\omega_\text{perception}$, which is illustrated in \cref{frame}.
An encoder, parameterized by convolutional neural networks (CNNs) and FCNs, is responsible for encoding information about objects and rewards into hidden representations.
Three FCNs use the hidden representations to predict the existence, the shape, and the coordinates of objects, which are explained below.

Objects can appear and disappear in many tasks.
Only the coordinates of present objects can be used by the symbolic policy, otherwise it will exploit the coordinates of missing objects and lose interpretability.
Therefore, we need to mask out the coordinates of missing objects during policy learning, which requires predicting the existence of objects.
One issue of existence prediction is that the distribution of objects can be long tailed.
For example, in the BeamRider task, the agent’s spaceship is present at almost every step, while torpedoes appear less frequently.
This issue is handled with the distribution-balanced focal loss~\citep{DistributionBalancedLoss}.

Specifically, for the $j$\textsuperscript{th} object in the $i$\textsuperscript{th} sample, let $c_{ij}=1$ if it is present, and $c_{ij}=0$ otherwise.
Let $C$ be the number of objects and $N$ be the number of samples.
This loss extend the focal loss~\citep{8237586} by weighting labels with their inverse frequency. It is given by 
\begin{equation}
\begin{aligned}
\mathcal{L}_\mathrm{exist} = \sum_{j=1}^{C}\bar\eta_{ij}\Big[ &c_{ij} (1 - p_{ij})^{\psi}  \log(p_{ij}) \\
+ (1-c_{ij})& p_{ij}^{\psi} \log(1 - p_{ij}) \Big],
\end{aligned}
\end{equation}
\noindent where $\psi$ is the modulating factor of the focal loss~\citep{8237586} and$p_{i,j}$ represents the probability generated by the existence layer. Details for computing $\bar{\eta}_{ij}$ is provided in \cref{db_focal}.

As for coordinate prediction, we use the L1 loss since the normalized coordinates can take small values.
Let $\mathbf{x}_i\in\mathbb{R}^{2C}$ be the vector for object coordinates in the $i$\textsuperscript{th} image and $\mathbf{\hat{x}}_i$ be its predicted value. For $j=1, 2, \dots, C$, $x_{i,2j}$ and $x_{i,2j+1}$ is the Y and X coordinate for the $j$\textsuperscript{th} object.
Then, the objective for coordinate prediction is given by:
\begin{equation}
\mathcal{L}_\mathrm{coor} = \sum_{j=1}^C c_{ij}\left[\left| x_{i,2j} - \hat{x}_{i,2j} \right| + \left| x_{i,2j+1} - \hat{x}_{i,2j+1} \right|\right].
\end{equation}
\noindent These losses are then averaged over all frames. Note that the predicted coordinates are clipped to $[0,1]$ when being used for policy learning, as they are supposed to be the coordinates of objects.
For shape prediction, the network is trained to predict the width and height of object bounding boxes. The loss, $\mathcal{L}_\mathrm{size}$, takes a similar form as $\mathcal{L}_\mathrm{coor}$, except that the prediction targets are replaced with the width and height of objects.
Overall, the objective for learning from the frame-symbol dataset is:
\begin{equation}\label{cnn_loss}
\mathcal{L}_\mathrm{cnn} =  \mathcal{L}_\mathrm{exist} + \mathcal{L}_\mathrm{coor}  + \mathcal{L}_\mathrm{size}. 
\end{equation}

\noindent Before policy learning, the perception module is pre-trained on frame-symbol datasets to equips agents with knowledge about objects.
We will verify this design choice in \cref{sec:experiments}.

\subsection{Learning Symbolic Policies}\label{subsec:end_to_end}

\textbf{EQL\quad}
We represent symbolic policies with the EQL network~\citep{martius2016extrapolation,sahoo2018learning}.
Given an input vector $\mathbf{h}^{(k-1)}\in\mathbb{R}^{d_{k-1}}$, its $k$\textsuperscript{th} layer first applies a transformation $\mathbf{g}^{(k)} =\mathbf{W}^{(k)} \mathbf{h}^{(k-1)} + \mathbf{b}^{(k)}$, where $\mathbf{W}^{(k)}\in\mathbb{R}^{d_k\times d_{k-1}}$ and $\mathbf{b}^{(k)}\in\mathbb{R}^{d_k}$ are learnable parameters.
$d_{k-1}$ and $d_k$ are the dimension of $\mathbf{h}^{(k-1)}$ and $\mathbf{g}^{(k)}$.

Compared to fully-connected layers (FCN), EQL has more flexible activation functions. 
The activation function for a vector $\mathbf{g}\in\mathbb{R}^{d_k}$, $\mathbf{f}: \mathbb{R}^{d_k}\to\mathbb{R}^{d'_k}$, is given by  
\begin{equation}\label{activation}
\mathbf{f}(\mathbf{g})=\left[\begin{array}{c}
    f_1(g_1) \\
    \vdots \\
    f_{u}(g_{u}) \\
    f_{u+1}(g_{u+1}, g_{u+2}) \\
    \vdots \\
    f_{d'_k}(g_{d_k-1}, g_{d_k})
\end{array}\right],
\end{equation}
where $f_1$, $f_2$, \dots, $f_{d'_k}\in \mathcal{F}$ are scalar functions.
$f_1, f_2, \dots, f_u$ are unary functions, and the remaining are binary functions. 
This flexible choice of activation functions enhances expressiveness of EQL.
In this work, $\mathcal{F}$ includes the square function, the cubic function, constants, the identity function, multiplication, and addition.

The final ingredient is sparsity regularization~\citep{martius2016extrapolation} required for deriving concise expressions, which imposes the following regularization to the parameters of a EQL network to avoid the singularity in the gradient as the weights go to 0.
\begin{equation}
\mathcal{L}_\text{reg}(w)= \begin{cases}|w|^{1 / 2} & |w| \geq a \\ \left(-\frac{w^4}{8 a^3}+\frac{3 w^2}{4 a}+\frac{3 a}{8}\right)^{1 / 2} & |w|<a\end{cases}
\end{equation}
Here, $a=0.05$ is a smoothing parameter.

\begin{algorithm}[t]
\small
\caption{Procedures for learning symbolic policy}
\label{algo}
\begin{algorithmic}
\STATE {\bfseries Input:} Pretrained perception module $\omega_\text{perception}$ and the frame-symbol dataset $\mathcal{D}_\text{symbol}$
\STATE {\bfseries Output: Fine-tuned perception module and the EQL actor $\pi_\text{EQL}$}
\STATE Initialize the neural actor $\pi_\text{nerual}$ and the EQL actor $\pi_\text{EQL}$ randomly.
\FOR{batch = $1,2,\ldots$}
    \STATE Collect online samples $\mathcal{D}_{\text{batch}}$ using $\pi_\text{nerual}$.
    \FOR{iteration=$1,2,...,n$}
        \IF{iteration==$n$}
            \STATE Update $\pi_\text{nerual}$, $\pi_\text{EQL}$, and $\omega_\text{perception}$ by minimizing $\mathcal{L}$ using $\mathcal{D}_{\text{batch}}$ and $\mathcal{D}_\text{symbol}$.
        \ELSE
            \STATE Update $\pi_\text{neural}$ and $\omega_\text{perception}$ by minimizing $\mathcal{L}_\text{ppo}$ using $\mathcal{D}_{\text{batch}}$.
        \ENDIF
    \ENDFOR
\ENDFOR
\end{algorithmic}
\end{algorithm}

\textbf{Neural Guidance\quad}We now discuss the challenges in learning symbolic policies.
The symbolic policy, referred to as the EQL actor $\pi_\text{EQL}$, computes action distributions using the predicted object coordinates.
While being intuitive, it is worth noting that the object coordinates $\mathbf{x}$ are subject to limited expressiveness.
Each element of $\mathbf{x}$ is bound to the Y or X coordinate of some objects, which means they are not distributed representations and cannot encode complex patterns as neural representations do.
What exacerbates the situation is that the expressions represented by $\pi_\text{EQL}$ are forced to be concise by $\mathcal{L}$.
In consequence, $\pi_\text{EQL}$ might not be able to represent \emph{all possible policies} and fail to explore well during policy learning.

We therefore opt for using $\pi_\text{EQL}$ to approximate only the optimal policy, which is less demanding than approximating all possible policies.
Inspired by~\citet{nguyen2021look,landajuela2021discovering}, we propose a neural guidance scheme that uses a neural actor $\pi_\text{neural}$ to interact with the environment.
$\pi_\text{neural}$ takes as input the hidden representations produced by the encoder of $\omega_\text{perception}$ and is not regularized by $\mathcal{L}_\text{0.5}$, so it can effectively explore the state-action space.
The EQL actor is trained to distill $\pi_\text{neural}$ using symbolic expressions and coordinate.
Compared to existing neural guidance schemes~\citep{nguyen2021look,landajuela2021discovering}, in our scheme the EQL actor and the neural actor are trained simultaneously rather than separately, which results in improved sample efficiency.

The procedures for learning symbolic policies are outlined in \cref{algo}.
Specifically, we optimize the neural actor using the Proximal Policy Optimization (PPO)~\citep{schulman2017proximal} algorithm, whose objective is denoted by $\mathcal{L}_\text{ppo}$.
As for the EQL actor, since we have access to $\pi_\text{neural}$, we minimize the cross entropy between the action distributions induced by the two actors, which is given by
\begin{equation}
\mathcal{L}_{\text{ng}}=-\mathbb{E}_{s,a\sim\pi_\text{neural}}\left[\log \left( \pi_\text{EQL}(a|s) \right)\right].
\end{equation}
The expectation is taken over online samples collected by $\pi_\text{neural}$.
For tasks with discrete action space, the expectation over actions can be calculated analytically.
To keep the accuracy of predicted coordinates, we also update $\omega_\text{perception}$ using $\mathcal{L}_\text{cnn}$.
In summary, the objective for learning $\pi_\text{EQL}$ and $\pi_\text{neural}$ jointly is given by
\begin{equation}
    \mathcal{L} = \mathcal{L}_\text{ppo} + \mathcal{L}_\text{ng} + \lambda_\text{reg}\mathcal{L}_\text{reg} + \lambda_\text{cnn}\mathcal{L}_\text{cnn}.
\end{equation}
where $\lambda_\text{reg},\lambda_\text{cnn}\in\mathbb{R}$ are hyper-parameters to balance policy learning, coordinate prediction, and sparsity regularization.

\textbf{Remarks\quad} The PPO algorithm reuses online samples for multiple iterations to improve sample efficiency.
In preliminary experiments, we found that optimizing $\mathcal{L}_\text{cnn}$ and $\mathcal{L}_\text{ng}$ for multiple iterations led to inferior performance, possibly due to drastic changes imposed on the perception module $\omega_\text{perception}$.
Thus, we optimize $\mathcal{L}$ only in the last iteration and $\mathcal{L}_\text{ppo}$ in all other iterations in \cref{algo}.

The perception module $\omega_\text{perception}$ sacrifices some auxiliary information for the sake of efficiency, since we only include the location, heights, and width in $\mathcal{D}_\text{symbol}$.
This is where end-to-end policy learning comes in.
By refining $\omega_\text{perception}$ using reward signals, the hidden layers of $\omega_\text{perception}$ can learn to capture features that are essential for task performance but overlooked during pre-training.
As shown by our results presented in \cref{cors} and \cref{methodablation}, end-to-end policy improves both task performance and coordinate prediction.



\newcolumntype{g}{>{\columncolor{lightgray}}c}
\begin{table*}[t!]
\centering
\scriptsize
\caption{\textbf{INSIGHT matches the performance of neural agents on all Atari tasks.} Performance of the proposed \ac{INSIGHT}, Neural (a neural agent that has the same network architecture as \ac{INSIGHT}), and existing NS-RL approaches (CGP~\citep{wilson2018evolving}, Diffses~\citep{zheng2022symbolic}, DSP~\citep{landajuela2021discovering}, and NUDGE~\citep{delfosse2023interpretable}) on nine Atari tasks.
\ac{INSIGHT} outperforms all NS-RL baselines and matches the performance of Neural for all nine tasks.}
\label{Performance}
\vskip 0.1in
\begin{threeparttable}
\resizebox{\textwidth}{!}{ 
\begin{tabular}{lcccccccg}
\toprule
Task & \ac{INSIGHT} & Neural & Coor-Neural &CGP\tnote{1} &DiffSES\tnote{2} &DSP\tnote{3} &NUDGE\tnote{4} & Human\tnote{5}\\ \midrule
Pong & $\textbf{20.9}\pm0.1$&$20.4\pm0.6$& $19.8\pm0$ &$20 \pm 0$&$20.2\pm 0$ &$-1 \pm 0$ &$-7.2\pm1.2$ &9.3    \\
BeamRider &  $3828.1\pm261.1$ &$3868.1\pm204.2$ & $1614.4\pm79.8$&$1341.6\pm 21$&/  &$354.6 \pm 20.3$ &/ &5775  \\
Enduro & $843.7\pm213.8$ &$676.9\pm730.4$ & $933.9\pm4.2$&$2\pm 0$ &/  &$41.1 \pm 15.5$ &$2.4\pm2.1$ &309.6  \\

Qbert &  
$16978.6\pm1936.1$   & $17879.2\pm1857.1$ & $13269.2\pm1024.4$&$770\pm94$&/  &$558.3 \pm 41.5$ &/ &13455  \\

SpaceInvaders & $\textbf{1232.6}\pm140.7$ &$1184.6\pm137.6$ & $717.7\pm15.3$&$1001\pm25$ &$792.4\pm 0$  &$222.6 \pm 9$ &$80\pm17$ &1652  \\

Seaquest & $\textbf{2665.7}\pm728.2$ &$1804.8\pm20.1$ & $1410.6\pm338.2$&$724\pm26$&/  &$193.3 \pm 12$ &$0\pm0$ &20182 \\
Breakout & $\textbf{409.6}\pm11.3$ &$259.6\pm183.8$ & $356.5\pm29.2$&$13.2\pm2$&/  &$4.3 \pm 0.5$ & $3.4\pm0.8$&31.8 \\

Freeway & $\textbf{32.7} \pm 0.1$ &$28.7\pm5.3$ & $32.4\pm0.1$&$28.2\pm0$&/  &$21.5 \pm 0.2$ &$21.4\pm0.8$ &29.6 \\
MsPacman & $\textbf{3042.5}\pm320.1$ &$2737.1\pm562.3$ & $2257.3\pm145.2$&$2568\pm72$4&/  &$2937.3 \pm 892.9$ &/ &15693  \\
\bottomrule
\end{tabular}
}
  \begin{tablenotes}
    \item[1] Results for CGP were taken from~\citep{wilson2018evolving}.
    \item[2] We used the results for Pong and SpaceInvaders reported by~\citet{zheng2022symbolic}, yet we were unable to obtain results for other tasks as the code is incomplete.
    \item[3] Since DSP cannot handle visual observations, we used pre-trained peception module of INSIGHT to extract object coordinates and used its released code for policy\\ learning.
    \item[4] We used the results reported by~\citet{delfosse2023interpretable} for Freeway and obtained results for Pong, Enduro, SpaceInvaders, Seaquest, and Breakout using its released code.\\ Due to the absence of predefined templates for all tasks within the codebase, we tailored the template originally designed for Asterix to fit the unique action spaces of the \\additional tasks. We were unable to obtain results for BeamRider, Qbert, and MsPacman since the ground-truth object locations are not available for them.
    \item[5] Results were taken from~\citep{wilson2018evolving}.
  \end{tablenotes}
\end{threeparttable}
\vskip -0.1in
\end{table*}

\subsection{Explaining Policies and Decisions}\label{enhanced-interpretability}
The pipeline for generating explanations starts with a step called concept grounding, which is followed by separate prompts for policy interpretations and decision explanations.
Full prompts and details are provided in \cref{Comprehensive Prompt Template Overview}.
Suppose a user who is familiar with a task wants to interpret a learned symbolic policy.
What can be tedious for the user is to associate concepts of the task, such as the goal and the influence of actions, with the construction of the symbolic policy.
For example, he or she has to find out which element in the coordinate vector corresponds to the location of a certain object.
We refer the process of establishing such a correspondence as \emph{concept grounding}.
Note that concept grounding is in fact not necessary for understanding explanations for policies and decisions.
For example, only knowledge of Pong is required to understand the conclusion part in the right part of \cref{gpt_analysis}, which are explanations for choosing action up at a particular state.
This observation inspires us to use LLMs to ground concepts and generate explanations, thereby reducing the user's intellectual burden.

Specifically, the prompt for concept grounding consists of \emph{task description} and \emph{policy description}.
The task description includes the goal of the task and the effects of actions, which is essential for steering the explanations toward task solving.
For Pong, it can be "There are two paddles in the screen...The agent earns a point if its opponent fails to strike the ball back...It needs to take one of the three actions: noop (take no operation)...".
As for policy description, we include the construction of the coordinate system and the expressions of the symbolic policy.
The semantics of coordinates are reflected with their names.
For example, the $x_{i,4}$ and $x_{i,5}$ for Pong are the X and Y coordinates of the agent's paddle.
To associate decisions with the location and motion of objects, we also instruct the LLM to infer the motion of objects from the coordinates in successive frames.

As shown in the upper left of \cref{gpt_analysis}, learned policies are formulated as a set of equations involving input variables (i.e. object coordinates), intermediate variables (e.g., $t_1$ in \cref{gpt_analysis}), and action logits.
Despite their simplicity, prompting LLMs with these equations only yields vacuous responses such as "the policy is complex, involving non-linear combinations of input variables".
Therefore, we apply the chain-of-thought principle and proceed in three steps: analyze the mapping from input variables to intermediate variables, analyze the mapping from intermediate variables to action logits, and finally summarize the findings.
Moreover, to mitigate issues such as hallucinations and uncontrollable outputs, we carefully design a set of rules for the LLM, as detailed in \cref{prompt_pong_policy}. For example, Rule 1 insists that explanations remain rooted in the policy's expressions, and rule 2 dictates that LLMs' policy explanations should be based on logits and action probabilities. In our preliminary experiments, these rules were able to improves the alignment between generated explanations and actual policy actions and makes outputs more precise and controllable.

To generate explanations for specific decisions, we provide the LLM with the exact values of object coordinates, the action being taken, and the gradients of action log-likelihood with respect to every object coordinate.
In addition, we instruct the LLM to assess the importance of input variables from the sensitivity perspective by mentioning the semantics of gradients, i.e. the sensitivity of action log-likelihood regarding to input variables, 
An example of generated explanations is shown in the bottom right of \cref{gpt_analysis}.

\section{Experiments}\label{sec:experiments}
This section evaluates the efficacy of INSIGHT for learning symbolic policy and structured state representations.
\cref{subsec:task_performance} reports the task performance of INSIGHT and validates our design choices, and \cref{subsec:interpretability} presents textual explanations for learned policies and specific decisions.

\subsection{Experimental Setup}
\textbf{Tasks\quad}
We consider the online RL setting and nine Atari tasks: Pong, BeamRider, Enduro, Qbert, SpaceInvaders, Seaquest, Breakout, Freeway, and MsPacman, which widely used in prior works~\citep{delfosse2023interpretable,landajuela2021discovering,zheng2022symbolic}. 
They range from simple motion control (e.g. Pong) to complex decision-making (e.g. Seaquest), and cover issues such as clean background (e.g. SpaceInvaders) vs noisy background (e.g. BeamRider) and constant object (e.g. Pong) vs varying objects (e.g. Enduro).
So with them we can examine INSIGHT comprehensively.
In the meantime, results on more realistic tasks are also important for validating NS-RL algorithms, so we also include results on MetaDrive \citep{li2022metadrive}, a challenging environment for autonomous driving.

\textbf{Evaluation Metrics\quad}
For task performance, we report the means and standard deviations of test returns after training agents for 10 million environment steps.
The higher, the better.
For coordinate prediction, we report the mean absolute error (MAE) of predicted coordinates, which is the result of dividing the total absolute error of coordinate prediction on a sample by the number of objects, the number of stacked frames, and two as we are predicting the X and Y coordinates.
It is within $[0,1]$, and the lower the better.
Since not all objects are relevant for symbolic policies, we include a variant of MAE, F-MAE, which only considers relevant objects.
All experiments are repeated for three seeds.

\color{blue}
\begin{table}[t!]
\centering
\scriptsize
\caption{\textbf{INSIGHT outperforms neural baselines on MetaDrive.} This table shows the performance of \ac{INSIGHT}, Neural and Random on MetaDrive when trained fro 1M, 2M, and 5M steps.}
\label{table:metadrive_performance_comparison}
\begin{threeparttable}
\begin{tabular}{lcccc}
\toprule
Success Rate & INSIGHT & Neural & Coor-Neural & Random \\ 
\midrule
1M & $\textbf{0.37}\pm0.21$ & $0.17\pm0.12$ & $0.07\pm0.05$ & $0\pm0$ \\
2M & $\textbf{0.42}\pm0.13$ & $0.19\pm0.06$ & $0.21\pm0.14$ & $0\pm0$ \\
5M & $\textbf{0.63}\pm0.11$ & $0.49\pm0.11$ & $0.41\pm0.12$ & $0\pm0$ \\
\bottomrule
\end{tabular}
\end{threeparttable}
\end{table}
\color{black}

\textbf{Baselines\quad}
For task performance, \ac{INSIGHT} is compared with Neural, DSP, Diffses, NUDGE, and CGP.
Neural is a deep RL alternative for \ac{INSIGHT} that uses the same network architecture but does not learn from the frame-symbol datasets.
The remaining four are approaches for NS-RL.
DSP~\citep{landajuela2021discovering} searches for symbolic policies from low-dimensional input using a recurrent neural network, while Diffses~\citep{zheng2022symbolic} uses frozen SPACE models to extract states and search for symbolic policies using genetic programming. 
NUDGE~\citep{delfosse2023interpretable} relies on the ground-truth structured states and represent policies with first-order differentiable logic.
Lastly, CGP~\citep{wilson2018evolving} leverages cartesian genetic programming for learning policies from images.

\textbf{\ac{INSIGHT} Variants\quad} 
To analyze our design choices, we include results for several variants of \ac{INSIGHT}. 
Specifically, w/o Pre-train is a variant that does not pre-train the perception model $\omega_\text{perception}$, and Fixed is a variant that freezes $\omega_\text{perception}$ in policy learning.
By w/o NG, we mean a variant that removes the neural guidance scheme and optimizes the EQL actor using $\mathcal{L}_\text{ppo}$ directly.
Meanwhile, Coor-Neural trains $\omega_\text{perception}$ and the neural actor $\pi_\text{neural}$ similarly with INSIGHT, but it uses the predicted coordinates as the input for $\pi_\text{neural}$.
SPACE-Neural extracts object coordinates from a pre-trained SPACE model, and SA-Neural utilizes the BO-QSA~\citep{jiaimproving}, a recent slot-attention algorithm, to extract latent representations of objects from images.

\textbf{Implementation Details\quad}
We use the standard pre-processing protocol for Atari tasks, which includes resizing, gray scaling, and frame stacking.
As for hyper-parameters, we use $0.001$ for $\lambda_\text{reg}$, 2 for $\lambda_\text{cnn}$.
We use the OpenAI's GPT-4 model~\citep{bubeck2023sparks} when generating explanations.
All the implementation details are provided in \cref{Architecture and Hyperparameters}.

\begin{table}[t]
\centering
\scriptsize
\caption{\textbf{INSIGHT is as efficient as Neural.} This table shows the inference time per state on Pong. The testing conditions are provided in \cref{Details of inference speed}. \ac{INSIGHT} is as efficient as Neural are is an order of magnitude faster than SA-Neural and SPACE-Neural.}
\label{timecompare}
\vskip 0.1in
\resizebox{0.47\textwidth}{!}{
\begin{tabular}{lccccc}
\toprule
Method  & INSIGHT & SPACE-Neural & SA-Neural & Neural & NUDGE   \\  \midrule
Time (ms) & $\textbf{2}$ & $50$ & $40$ & $\textbf{2}$ & $60$ \\
\bottomrule
\end{tabular}}
\vskip -0.1in
\end{table}

\begin{figure}[t]
  \centering
  \includegraphics[width=0.9\linewidth]{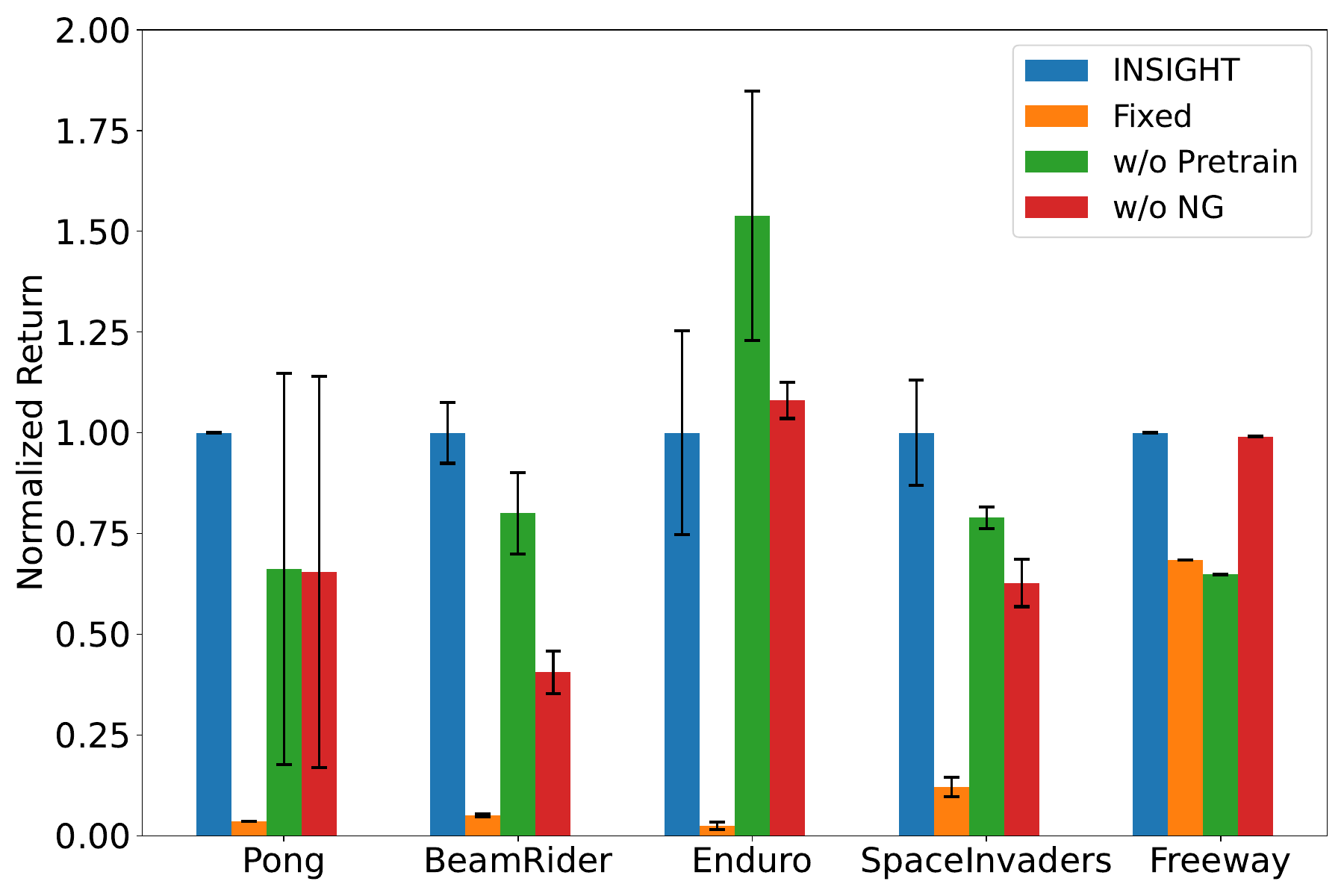}
    \caption{\textbf{Each component of INSIGHT is critical for overall performance.  }Detailed performance analyses of \ac{INSIGHT} and its variants across five tasks are presented. Refer to \cref{remain_methodablation} for results for the remaining four tasks. Test returns are normalized so that \ac{INSIGHT} corresponds to one and random policy is zero. Fixing the perception module during policy learning (i.e., Fixed) hinders performance for all tasks, indicating that it is crucial to refine states with reward signals. The results of w/o Pretrain and w/o NG show that pre-training the perception module and the proposed neural guidance scheme also improves performance.}
\label{methodablation}
\vskip -0.1in
\end{figure}

\begin{table*}[t!]
\centering
\scriptsize
\caption{\textbf{INSIGHT significantly enhances coordinate prediction for policy-relevant objects through policy learning.} MAE and F-MAE measure the coordinate prediction error, with lower values indicating better performance. F-MAE measures the mean absolute error of coordinate prediction for policy-related objects. Numbers are multiplied by 100. Refer to \cref{cors-remain} for results for the other four tasks. By comparing the results of INSIGHT with those of Fixed, we see that refining state representations with rewards improves coordinate prediction for policy-related objects. The proposed neural guidance scheme is crucial for such improvement, as implied by results of w/o NG.}
\label{cors}
\vskip 0.1in
\resizebox{\textwidth}{!}{ 
\begin{tabular}{lcccccccccc}
\toprule
 &\multicolumn{2}{c}{INSIGHT}             &\multicolumn{2}{c}{w/o Pretrain}  &\multicolumn{2}{c}{Fixed}           &\multicolumn{2}{c}{w/o NG}     &\multicolumn{2}{c}{Coor-Neural}   \\ 
Tasks &MAE &F-MAE &MAE &F-MAE &MAE &F-MAE &MAE &F-MAE &MAE &F-MAE \\ \midrule
Pong            &$1.4\pm0.9$ &$\textbf{1.3}\pm0.9$       &$2.9\pm3.1$ &$3.5\pm3.8$      &$2.8\pm0.5$ &$3.1\pm0.3$  &$1.7\pm0.9$  &$1.6\pm0.9$  &$\textbf{0.9}\pm 0$ &/     \\
BeamRider       &$\textbf{5.5}\pm0$ &$\textbf{3.2}\pm0.3$    &$5.8\pm0.1$ &$3.5\pm0.1$      &$\textbf{5.5}\pm0.4$ &$4.9\pm0.1$     &$6.2\pm0.1$&$5.9\pm0$  &$5.7\pm 0$  &/     \\
Enduro          &$13.5\pm0$ &$\textbf{6.7}\pm0.3$     &$13.1\pm0.1$&$8.2\pm0.2$       &$\textbf{12.7}\pm0.5$   &$12.2\pm0.5$    &$13.4\pm0.2$ &$13.4\pm0.2$  &$13.7\pm 0.1$    &/       \\
SpaceInvaders   &$10.3\pm0.4$ &$8.1\pm0.9$      &$10.6\pm0.2$&$\textbf{6.5}\pm0.6$       &$\textbf{9.9}\pm0.5$   &$11.1\pm0.4$    &$11.1\pm0.3$ &$11.1\pm0.4$  &$10.3\pm 0.3$    &/        \\ 
Freeway         &$16.3\pm0.3$ &$\textbf{14.8}\pm1.6$         &$23.4\pm0$  &$19.4\pm5.6$       &$\textbf{15.7}\pm0.2$   &$18.7\pm0.8$    &$16.4\pm0.2$ &$16.4\pm0.2$  &$16.4\pm 0.2$    &/      \\
\midrule
\end{tabular}}
\vskip -0.1in
\end{table*}

\subsection{Quantitative Results}\label{subsec:task_performance}
\textbf{Improved Task Performance\quad}
\Cref{Performance} shows that except for Pong, where visual observations are less involved, NS-RL baselines are outperformed by Neural by a wide margin.
In contrast, \ac{INSIGHT} can match or even outperform Neural.
In addition, in \cref{Performance-cleanrl} we show that our implementation for Neural matches the performance of CleanRL~\citep{huang2022cleanrl}, an open-source implementation for deep RL agents.
\cref{table:metadrive_performance_comparison} shows the success rates of INSIGHT and neural baselines over 1M, 2M, and 5M timesteps for Metadrive. INSIGHT outperforms the neural baselines at all timesteps.
These results highlight that INSIGHT, by learning structured states and symbolic policies jointly, overcomes the performance drawback of existing NS-RL approaches.
We now present an ablation analysis to explain the improvement.

\textbf{Better Efficiency\quad} 
As mentioned in \cref{subsec:state_learning}, the perception module of INSIGHT is designed to be efficient.
\Cref{timecompare} shows that INSIGHT is an order of magnitude faster than both SPACE-Neural and SA-Neural, which verifies this design.
In addition, \cref{sa} shows that agents learn faster when learning from $\mathcal{D}_\text{symbol}$.
A possible reason is that $\mathcal{D}_\text{symbol}$ provides direct supervision for object locations. When trained with image reconstruction objectives, agents are not forced to focused on individual objects.

\begin{figure}[t]
  \centering
  \includegraphics[width=0.8\linewidth]{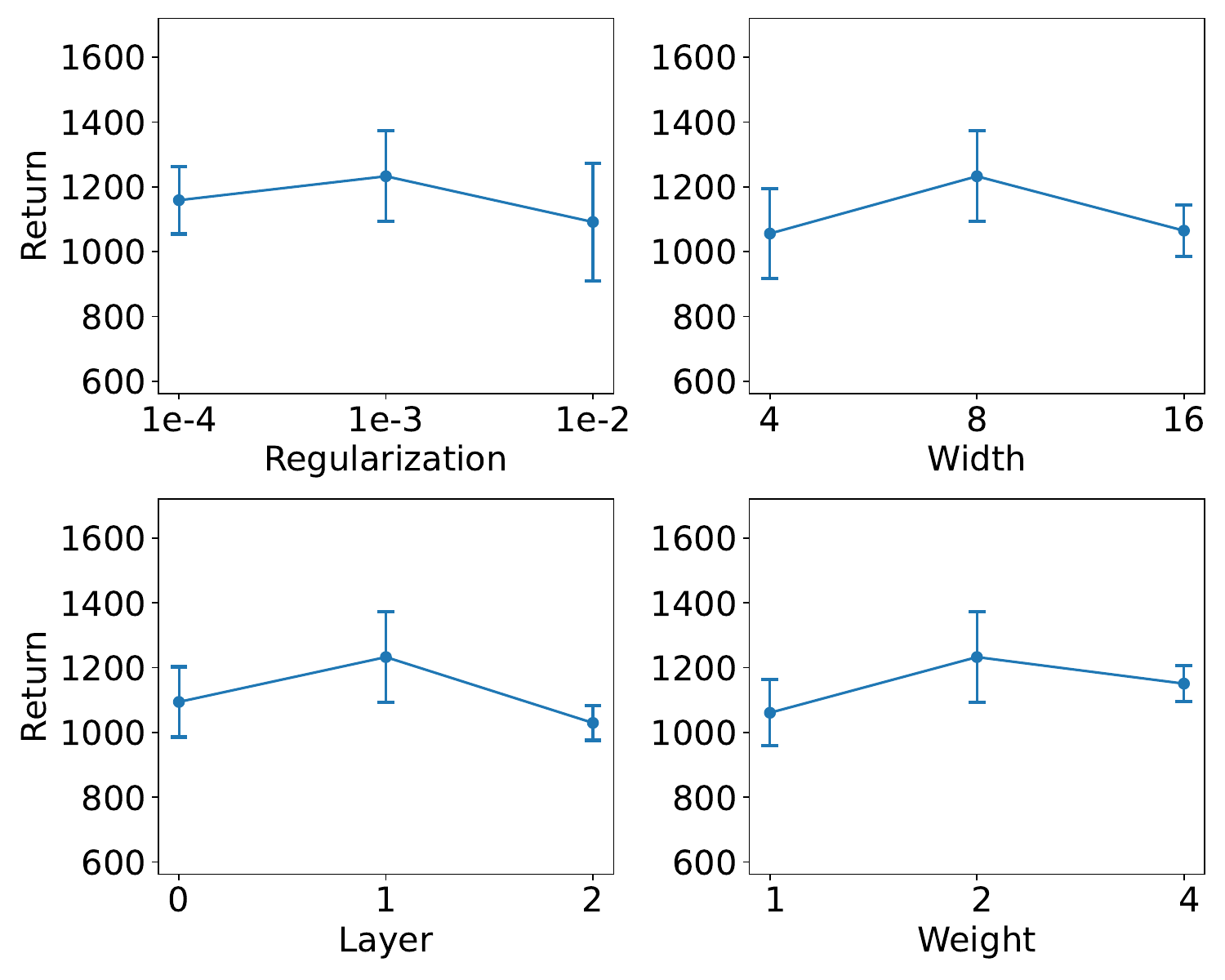}
    \caption{\textbf{INSIGHT demonstrates robustness to hyper-parameters.} Examining the influence of sparsity regularization weight $\lambda_\text{reg}$, the EQL actor's width/layers, and the weight of $\mathcal{L}_\text{cnn}$ on SpaceInvaders performance. Results for additional tasks are available in \cref{hyper ablation remain}. Overall, INSIGHT shows substantial robustness to variations in hyper-parameters.}
\label{hyperablation}
\vskip -0.1in
\end{figure}

\textbf{Benefits of End-to-End Policy Learning\quad}
We now examine our claim for refining structured states with both visual and rewards from the perspective of task performance and coordinate prediction.
\cref{methodablation} shows that when compared to the variant Fixed, INSIGHT has significant better performance for all five task, demonstrating that INSIGHT's performance is largely determined by its ability to refine structured states with reward signals.
\cref{cors} further reveals a clue for the performance improvement.
Compared to Fixed, \ac{INSIGHT} has higher MAE for four tasks but lower F-MAE for all five tasks.
In addition, readers may refer to \cref{Coordinates before and after training} for visualizations of predicted coordinates.
These findings suggest that reward signals can guide agents to improve the coordinate prediction of policy-relevant objects.

\begin{figure*}[ht]
     \centering 
     \begin{subfigure}[c]{0.49\textwidth}
         \centering
         \includegraphics[width=\textwidth]{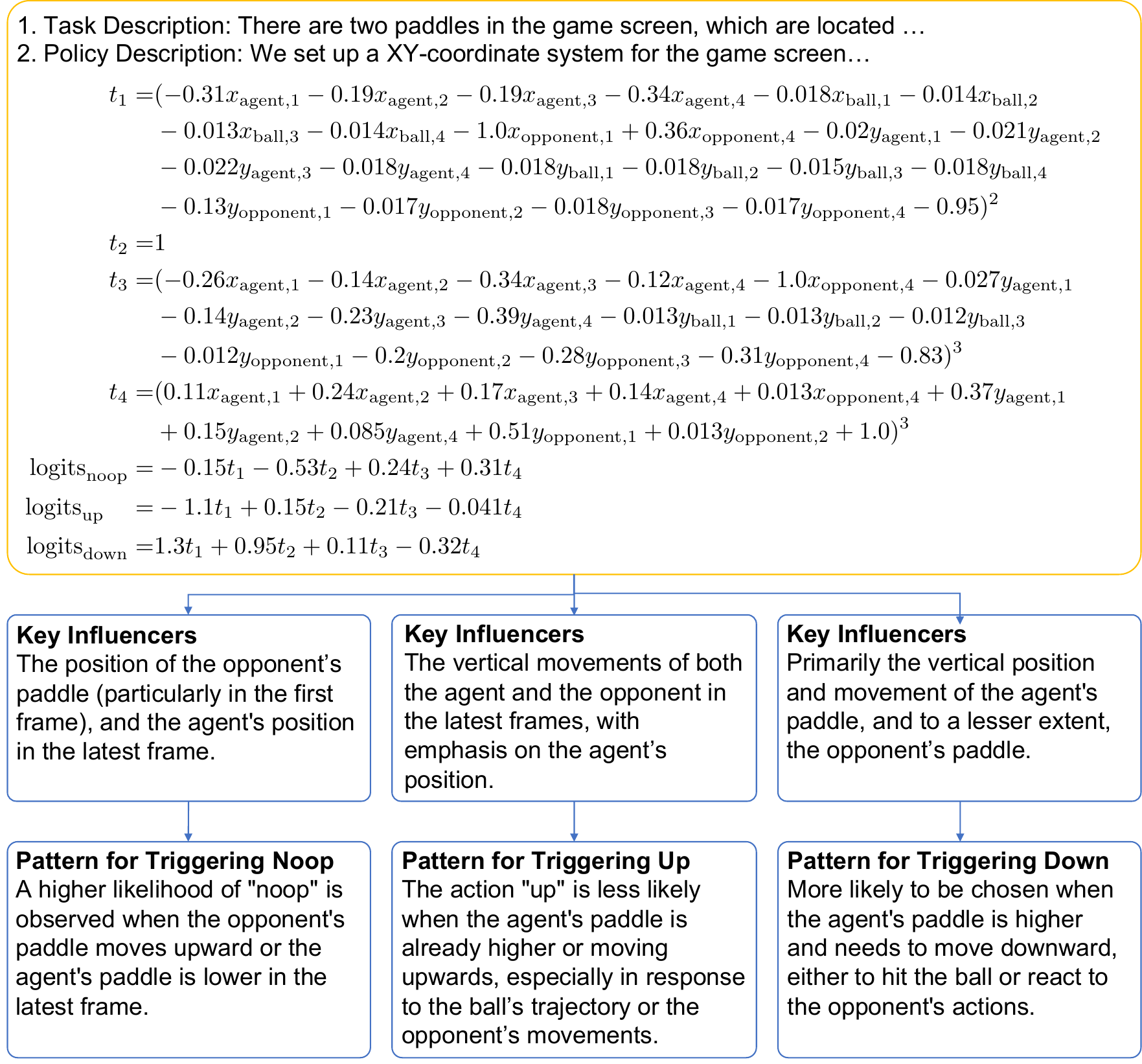}
     \end{subfigure}
     \hfill
     \begin{subfigure}[c]{0.49\textwidth}
         \centering
         \includegraphics[width=\textwidth]{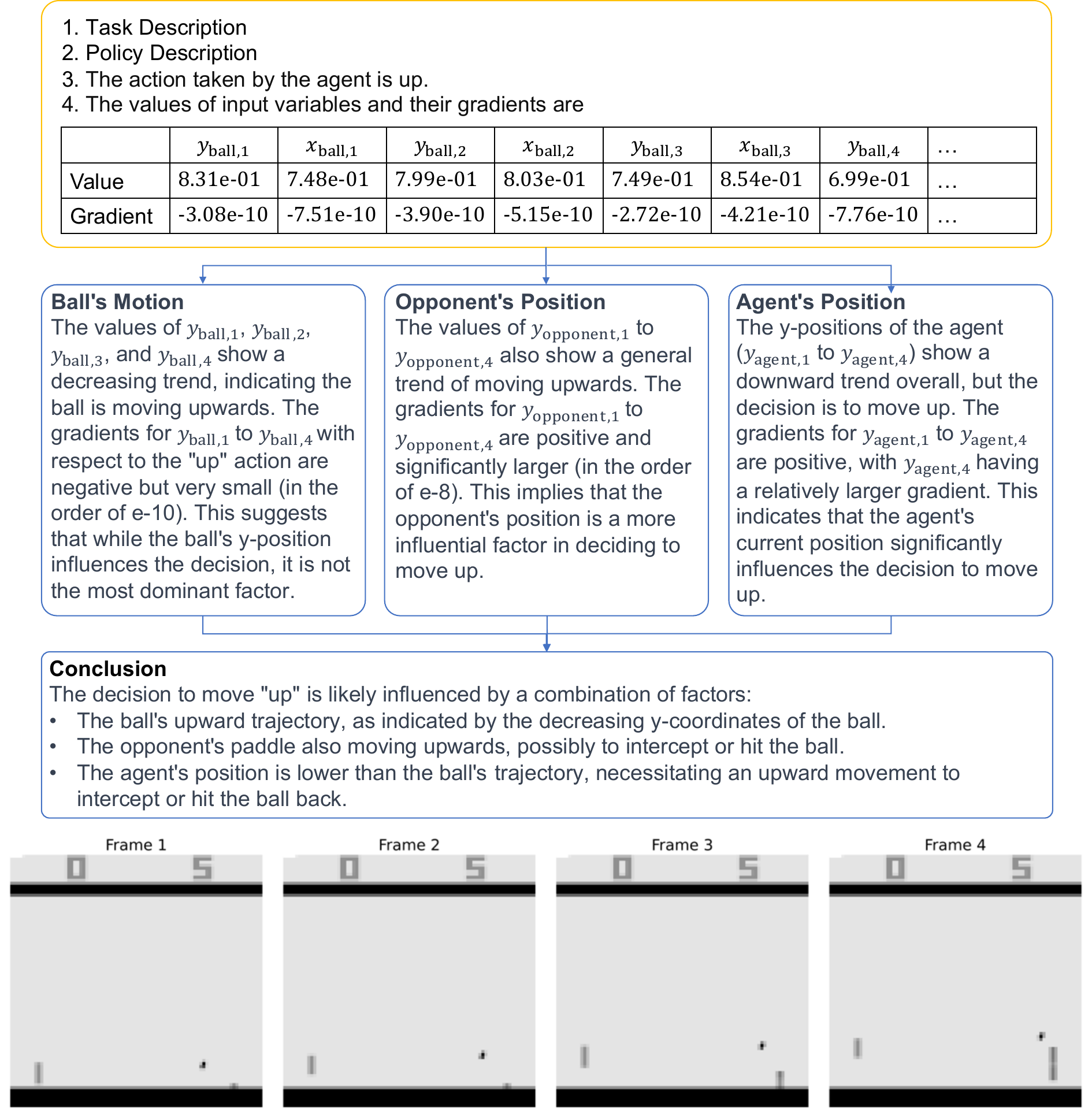}
     \end{subfigure}
     \hfill
        \caption{Examples for textual explanations for Pong. Readers may refer to \cref{Comprehensive Prompt Template Overview} for full prompts. \textbf{Left}: interpretations for a learned policy. The interpretations identify influential input variables and summarize triggering patterns of actions. \textbf{Right}: explanations for an action taken at a state. The four images located at the bottom illustrate the state. The motion of the ball and the opponent's paddle are deduced from input variables, which are used for supporting explanations of actions.}
        \label{gpt_analysis}
    \vskip -0.1in
\end{figure*}

\textbf{Benefits of Pre-Training the Perception Module\quad} In \cref{subsec:state_learning}, we suggest pre-training the perception module before policy learning.
\Cref{methodablation} shows that the pre-training step indeed improves performance for four tasks, and \cref{cors} indicates that the improvement may be the result of better prediction for policy-relevant objects. Despite the imperfections and redundancy in the representation space obtained through pre-training, it still offers valuable information about policy-related objects, serving as a strong initialization for subsequent end-to-end training. 
Meanwhile, this extra step does complicate the training protocol of \ac{INSIGHT}.
One possible improvement is to use off-policy algorithms for policy learning, which allow the perception module to be optimized for more gradient steps.

\textbf{
Efficacy of the Proposed Neural Guidance Scheme\quad}
We introduce the neural guidance scheme to address the limited expressiveness of object coordinates.
The effectiveness of this scheme is supported by results in \Cref{methodablation}–w/o NG performs worse than INSIGHT for three tasks out of five tasks.
Moreover, Coor-Neural is outperformed by Neural for six tasks in \cref{Performance}, suggesting that the object coordinates are not sufficiently expressive for online policy learning.
\cref{cors} reveals that unlike the case of INSIGHT, the F-MAE and MAE of w/o NG are almost the same, implying that w/o NG cannot improve coordinate prediction using reward signals.
Thus, the neural guidance scheme plays a key role in refining coordinate prediction with rewards. 

\textbf{Robustness Against Hyper-Parameters\quad}
Lastly, \cref{hyperablation} shows the influence of four hyper-parameters for SpaceInvaders, which indicates that the performance of INSIGHT are quite robust against the hyper-parameters.
Results for other tasks are provided in \cref{hyper ablation remain}.
Since our goal is to illustrate the efficacy of INSIGHT rather than benchmarking, we use the same values for hyper-parameters for all tasks.

In summary, the ability to refine states with reward signals is decisive for performance improvement, possibly due to the improved coordinate prediction for policy-relevant objects.
The proposed neural guidance scheme also plays a role in improving task performance and coordinate prediction.

\subsection{Textual Explanations}\label{subsec:interpretability}

The left part of \cref{gpt_analysis} presents interpretations of a policy learned for Pong, which requires an agent to control the right paddle to hit the ball to the left-hand side. The agent earns a point if the opponent fails to hit the ball back.
Both paddles move only vertically.

The bottom left part of \cref{gpt_analysis} shows policy interpretations generated by INSIGHT.
It shows that the influential variables are correctly identified.
For example, $y_\text{opponent,1}$ is recognized as influential for action noop due to its coefficient (-0.13), and the coordinates of the ball are omitted for their small coefficients. 
However, some triggering patterns are less convincing.
Action up is considered to be less likely when the agent's paddle moves upward ($y_\text{agent}$ decreases).
While this is correct when considering $t_3$, it is in fact difficult to determine how $t_1$ changes in response to decrease of $y_\text{agent}$.
These observations confirms the LLM's ability to perform step-by-step analysis for symbolic policies.

The right part of \cref{gpt_analysis} showcases decision explanations. 
While the movements of the ball is correctly deduced, there is a mistake for the agent's paddle, possibly due to fluctuations in the coordinate predictions.
The generated explanations reveals that the decision is more sensitive to the position of the opponent's paddle than the ball, suggesting that the agent is exploiting the opponent's fixed policy.
Similar phenomena are also observed in recent NS-RL works~\citep{delfosse2024interpretable}, which highlights the importance of inspecting learned policies manually.
Nevertheless, INSIGHT can generate credible explanations by associating interpretations with facts.

Overall, despite the minor mistakes, both the policy interpretation and the decision explanation are friendly to non-expert users and reveal some patterns in the agent's decision-making process.

\section{Conclusions}
We propose \ac{INSIGHT}, a framework that uses object coordinates as structured state representations and learns symbolic policies from visual input.
INSIGHT is able to refine the structured states with reward signals by distilling vision foundation models into a scalable perception module, and it leverages a new neural guidance scheme to learn competitive symbolic policies from object coordinates, thereby overcoming the performance bottleneck of previous NS-RL approaches.
Moreover, to improve model transparency for non-expert users, INSIGHT can generate language explanations for learned policies and specific decisions with LLMs. 
With experiments on nine Atari tasks and a MetaDrive task, we show that INSIGHT outperforms all NS-RL baselines and further reveal that the improvement can be explained by the improved coordinate prediction for policy-relevant objects.
We also showcase language explanations for learned policies and decisions.

\paragraph{Limitations} Currently, the EQL network in \ac{INSIGHT} cannot express logical operations required by some reasoning tasks.
Furthermore, quantitative evaluation of the explanations is also an interesting future topic.

\section*{Acknowledgments}
C. Fang was supported by the National Science and Technology Major Project (2022ZD0114902), the NSF China (No. 62376008). 
 The authors would like to thank four anonymous reviewers for constructive feedback.

\section*{Impact Statement}
This paper aims to improve RL's transparency without sacrificing task performance, which paves the road towards trustworthy RL applications.
From a societal point of view, we aim to reduce the barrier between the general public and RL models, thus advancing the broad application of RL agents.

\bibliographystyle{icml2024}
\bibliography{reference_header,references}

\clearpage
\appendix

\renewcommand\thefigure{A\arabic{figure}}
\setcounter{figure}{0}
\renewcommand\thetable{A\arabic{table}}
\setcounter{table}{0}
\renewcommand\theequation{A\arabic{equation}}
\setcounter{equation}{0}
\pagenumbering{arabic}
\renewcommand*{\thepage}{A\arabic{page}}
\setcounter{footnote}{0}
\resetlinenumber[1]

\appendix
\onecolumn
\section{Details of INSIGHT}
\subsection{Details of the Frame-Symbol Dataset}\label{Enhanced Description of Frame-Symbol Dataset Generation}

This section provides a detailed explanation of the frame-symbol dataset generation process, expanding on the preliminary overview presented in \cref{subsec:state_learning}.

The generation of the dataset begins with a comprehensive training regimen of 10 million steps using the neural baseline. This regimen incorporates a neural policy tailored for interaction with the environment. During the final 1 million steps, specifically starting from the 9-millionth step, we captured 10,000 consecutive frames spanning multiple episodes, thereby forming an unsupervised frame dataset. This phase primarily focuses on documenting environmental interactions via image captures, with a special emphasis on acquiring objects in high-reward scenarios.

Following this, the initial frame undergoes processing through FastSAM, which identifies and segregates unique objects by extracting their masks. Without assuming prior knowledge of the number of objects in the environment. To accommodate various environments, we establish a maximum object count of 256 for all experiments, a threshold that is sufficient for all Atari games. The objects are ordered based on their confidence scores from FastSAM. These masks serve as inputs for the DeAot module, enabling object tracking across the subsequent 9,999 frames. To maintain the consistency of object IDs, these 10,000 frames were spliced into a whole video and read by DeAot. FastSAM re-evaluates the scene every tenth frame to include new objects, leading to an increase in the object count. This increment prompts DeAot to start tracking these newly identified objects. For improved segmentation and tracking, all frames are resized to a resolution of 1024×1024 pixels.

During the segmentation phase using FastSAM, objects are excluded if their confidence level does not meet or exceed a threshold of 0.9. This stringent selection criterion is pivotal for minimizing misidentifications attributable to environmental factors. Moreover, within the same frame, a distinction is made between connected and non-connected masks. Connected masks that do not overlap with previously tracked objects are deemed new. Conversely, objects that overlap with an existing mask by 50\% or less are also classified as new, i.e., IOU=0.5. The rationale for the former is to avoid fragmenting a single object into multiple smaller segments as much as possible, and for the latter, it is because the tracking module tends to recognize non-connected, similar objects as a single entity, necessitating their re-identification and enumeration.

In the dataset's final development phase, the dataset was re-segmented starting at $0^{th}$ frame and the FastSAM module is deactivated, allowing DeAot to autonomously track objects from the initial to the final frame. The resulting dataset encompasses detailed information like object coordinates, bounding boxes, and RGB values, linked to their corresponding frames. When an object is missing from a frame, it is noted as 'non-existent' in the dataset, and its coordinates are set to (0,0). This approach allows for a systematic way to account for absent objects without disrupting the overall tracking and identification process.
Ultimately, this leads to the formation of a comprehensive frame-symbol dataset, where frames are stored at a resolution of 512×512 pixels. This process mitigates a previously identified limitation where the tracking model struggled to consistently detect objects in each frame. Conforming to the standard supervised training approach, the dataset is divided into training and test sets in an 80:20 ratio.

\subsection{Label Weights of the Distribution-Balanced Focal Loss}\label{db_focal}
Denote by $n_j={1}/{\sum_{i=1}^N c_{ij}}$ the inverse frequency of the $j$\textsuperscript{th} label, where $N$ is the number of samples in $\mathcal{D}_\text{symbol}$.
The inverse frequency is further normalized, since the number of labels varies across samples, and transformed to the range $[\alpha, \alpha+1]$ for numerical stability.
That is, the weight of label $c_{ij}$ is given by $\bar\eta_{ij}=\alpha+\sigma(\beta (\eta_{ij}-\mu))$, where $\eta_{ij}={n_{j}}/{\sum_{j'}^Cc_{ij'}n_{j'}}$ and $\sigma(\cdot)$ is the sigmoid function.
$\alpha$, $\beta$, and $\mu$ are hyper-parameters.

\subsection{Details of F-MAE}\label{Details of F-MAE}

The F-MAE metric evaluates the precision of predicted object coordinates within frames of Atari tasks, with a focus on objects essential for the agent's decision-making. Unlike traditional MAE, F-MAE targets a subset of objects identified as critical through symbolic regression. For a dataset $\mathcal{D}_\text{symbol}$ containing $N$ samples, the presence of the $j^\text{th}$ object in the $i^\text{th}$ sample is marked by $c_{ij}$, where $c_{ij}=1$ indicates the object's presence, and $c_{ij}=0$ its absence. Let $\mathbf{x}_i\in\mathbb{R}^{2C}$ represent the vector of object coordinates in the $i^\text{th}$ image, and $\mathbf{\hat{x}}_i$ its predicted counterpart. For objects numbered $j=1, 2, \dots, C$, the coordinates $x_{i,2j}$ and $x_{i,2j+1}$ correspond to the Y and X positions, respectively. The F-MAE, focusing on a critical subset of objects denoted as $S$, is calculated using \cref{F-MAE formular}:

\begin{align}\label{F-MAE formular}
\text{F-MAE} = & \frac{1}{2EN|S|} \sum_{i=1}^N\sum_{j=1}^C s_{ij}c_{ij}\left(\left| x_{i,2j} - \hat{x}_{i,2j} \right| + \left| x_{i,2j+1} - \hat{x}_{i,2j+1} \right|\right).
\end{align}

Here, $s_{ij}$ denotes the inclusion of the $j^\text{th}$ object in the $i^\text{th}$ sample within the filtered subset, taking a value of 1 when included and 0 otherwise. The term $E$ represents the total number of frames featuring the objects after filtering. The division by 2 accounts for the mean impact of the x and y coordinates. The term $|S|$ signifies the count of objects post-filtering, with the division by $|S|$ adjusting for the impact of individual objects. This formula constrains the F-MAE's range between 0 and 1, where 0 indicates perfect accuracy and 1 denotes complete inaccuracy, thus providing a clear metric for evaluating object coordinate prediction precision within task frames.

\subsection{Conditions for Testing Inference Speed}\label{Details of inference speed}

In the experiments conducted as detailed in \cref{timecompare}, the hardware setup comprised an AMD Ryzen 9 5950X 16-Core Processor for CPU, an NVIDIA GeForce RTX 3090 Ti as the graphics card, and 24564MiB of video memory. Each experiment involved executing 1000 steps on the Pong task, from which the average single-step inference time of the model was calculated.

\section{Experimental Setup}
\label{Experimental Setting}
\subsection{Architecture and Hyperparameters}
\label{Architecture and Hyperparameters}
\paragraph{CNN Encoder Structure}

In \cref{tab:CNN encoder}, the structure of the CNN encoder is elaborated. Comprising three convolutional layers, each layer is distinctively configured with varying kernel sizes, strides, padding, and channel outputs. The initial resolution of the image is defined as 84x84 pixels, encompassing four channels. These channels amalgamate the grayscale images from the previous four temporal frames, capturing motion information.

\begin{table*}[t!]
\centering
\caption{Hyperparameters of the CNN encoder.}
\label{tab:CNN encoder}
\begin{tabular}{cccccc}
\toprule
Hyperparameter &  Value \\
\midrule
Resolution   & 84x84  \\
Image Channels  & 4  \\
\toprule
Encoder Configuration \\
\toprule
Layer & Kernel Size & Stride & Padding & Channels & Activation \\
\midrule
Conv1 & 5x5 & 2 & 2 & 32 & ReLU \\
Conv2 & 5x5 & 2 & 2 & 64 & ReLU \\
Conv3 & 5x5 & 1 & 2 & 64 & ReLU \\
\toprule
Post-Convolution Layers \\
\toprule
Layer & Out Feature & Activation  \\
\midrule
Flatten & - & - \\
Linear & 2048 & ReLU  \\
LayerNorm & 2048 & - \\
\toprule
Output Layers \\
\toprule
Layer & Structure & Activation  \\
\midrule
Existence Layer & Linear(2048, 1024) & ReLU \\
                 & Linear(1024, 1024) & - \\
Coordinate Layer & Linear(2048, 2048) & ReLU \\
                 & Linear(2048, 2048) & - \\
Shape Layer & Linear(2048, 512) & ReLU \\
            & Linear(512, 512) & - \\
\bottomrule
\end{tabular}
\end{table*}

The initial convolutional layer (Conv1) utilizes a 5x5 kernel, a stride of 2, and padding of 2 to output 32 channels. This configuration begins the feature extraction process, reducing spatial dimensions while enriching the feature map's depth. The following layers (Conv2 and Conv3) maintain this configuration but with an increased channel output of 64, capturing more intricate features.

Post-convolution, the network integrates a flattening step, converting the multi-dimensional feature maps into a singular vector. This vector feeds into a fully connected linear layer with 2048 output features and ReLU activation, transforming the detailed convolutional features for subsequent analysis.

A layer normalization follows the first linear layer, enhancing the learning process's stability and efficiency. This normalization standardizes data scales and enables faster training through higher learning rates.

In the final stage, the network includes distinct output layers for existence, coordinate, and shape predictions. The Existence Layer initially compresses the feature dimensions from 2048 to 1024 using a ReLU activation function. This is succeeded by a subsequent linear layer, which maintains this reduced feature dimension. The Coordinate Layer, conversely, preserves the feature count at 2048, while the Shape Layer diminishes it to 512. Both these layers incorporate ReLU activations and are followed by linear transformations for processing.

\paragraph{EQL Structure}

The input dimension of the EQL network is configured as 2048, specifically designed to accommodate the coordinate representation generated by the encoder. These coordinates are initially processed through a hidden layer, followed by a custom activation function. The activation function employs a variety of operations to avoid excessive gradient explosion and ensure sufficient representation ability. These include squaring (\(f(x) = x^2\)), cubing (\(f(x) = x^3\)), constant (\(f(x) = c\)), identity (\(f(x) = x\)), product (\(f(x, y) = xy\)), and addition (\(f(x, y) = x + y\)). Each of these functions is iteratively applied four times within the hidden layer, as detailed in \cref{tab:eql_hyperparameters}. Subsequently, the output layer generates a symbolic expression corresponding to each action dimension within the environment. This process includes a transition through a softmax layer, where the final action is derived by random sampling from the resulting probability distribution. Furthermore, the output of the EQL network is multiplied by a temperature coefficient $t_{eql} = 10$, enhancing sensitivity to coordinate changes and aiding in the preservation of object coordinate prediction during policy learning.

\begin{table}[t!]
\centering
\caption{Hyperparameters of EQL network.}
\label{tab:eql_hyperparameters}
\begin{tabular}{cc}
\toprule
\textbf{Hyperparameter}                            & \textbf{Value}                      \\
\midrule
Input Dimensions                                   & 2048                                \\
Activation Function 1                    & \( f(x) = x^2 \)                    \\
Activation Function 2                    & \( f(x) = x^3 \)                    \\
Activation Function 3                   & \( f(x) = c \)                      \\
Activation Function 4                    & \( f(x) = x \)                      \\
Activation Function 5                   & \( f(x, y) = xy \)                  \\
Activation Function 6                       & \( f(x, y) = x + y \)               \\
Number of Repetitions  & 4                             \\
Number of Hidden Layers  & 1                             \\
$t_{eql}$  & 10                             \\
\bottomrule
\end{tabular}
\end{table}

\begin{table}[t!]
\centering
\caption{Hyperparameter of pretraining.}
\label{tab:pretrain}
\begin{tabular}{ccc}
\toprule
\textbf{Hyperparameter} & \textbf{Value}  \\
\midrule
Epoch & 600 \\
Batch Size & 32 \\
Learning Rate & \(3 \times 10^{-4}\)  \\
Weight Decay & \(1 \times 10^{-4}\) \\
$\alpha$ of $\mathcal{L}_\text{exist}$ & 0.1 \\
$\beta$ of $\mathcal{L}_\text{exist}$ & 10 \\
$\mu$ of $\mathcal{L}_\text{exist}$ & Mean Value of $n_j$ \\
$\gamma$ of $\mathcal{L}_\text{exist}$ & 2 \\
Optimizer & Adam \\
Loss Function & \cref{cnn_loss}\\
\bottomrule
\end{tabular}
\end{table}

\begin{table}[t!]
\centering
\caption{Hyperparameters of Policy Learning.}
\label{tab:policy_learning}
\begin{tabular}{cc}
\toprule
\textbf{Hyperparameter} & \textbf{Value} \\
\midrule
Total Steps & 10M \\
Learning Rate & \(2.5 \times 10^{-4}\) \\
Batch Size & 1024 \\
Initial value of $\lambda_\text{reg}$ & \(1 \times 10^{-3}\) \\
Coefficient for annealing $\lambda_\text{reg}$ linearly & \(\frac{\text{Update}-1}{\text{Total Updates}}\) \\
$\lambda_\text{cnn}$ & 2 \\
\bottomrule
\end{tabular}
\end{table}

\paragraph{Pretraining}
The pretraining stage employed the loss function detailed in \cref{cnn_loss}, spanning 600 epochs with a batch size of 32. To counteract overfitting, a weight decay regularization of \(1 \times 10^{-4}\) was applied. The learning rate was established at \(3 \times 10^{-4}\), with the Adam optimizer selected for its effectiveness in gradient descent optimization. Detailed specifics of the parameters, including their command line arguments and values, are provided in \cref{tab:pretrain}.

\paragraph{Policy Learing}

In the policy learning phase, following the common settings of Atari tasks, INSIGHT interacts with the environment for 10M steps. PPO with a learning rate of 2.5e-4 uses the collected rewards to optimize the model. The batchsize of each update is 1024. In order to ensure that the network can learn useful policies while ensuring the coefficient of the policy, we choose to use 1e-3 regularization and increase the regularization coefficient from 0 to 1 in each update. In addition, we multiply the coefficient of \cref{cnn_loss} by 2 to enhance the accuracy of coordinate prediction. All detailed parameters are summarized in \cref{tab:policy_learning}.

\section{Additional Experimental Results}\label{Additional experimental results}

This section encompasses all supplementary experiments, providing a comprehensive overview.

\subsection{Comprehensive Evaluation of Representations}\label{Comprehensive Evaluation of Representations}

We highlight an additional advantage of adopting a distilled vision-based model: enhanced efficiency in task learning through improved representations. \Cref{sa} presents the task performance of different approach to extract state representations accompanied by neural actors.

\begin{table*}[t!]
\centering
\scriptsize
\caption{\textbf{Distilled structured representations significantly enhance performance.} Evaluating the performance of SA-Neural, SPACE-Neural, Coor-Neural, and Neural after training for one million steps. Coor-Neural surpasses both SA-Neural and SPACE-Neural in most tasks and stands as the only competitor to Neural. This demonstrates the accelerated learning achieved by agents using our proposed method for structured state representation.}
\label{sa}
\begin{tabular}{lcccc}
\toprule
Tasks & SA-Neural & SPACE-Neural  & Coor-Neural & Neural \\ \midrule
Pong & $5.4\pm 0.5$&  $-20.6
 \pm 0$ &$10.4\pm 6.2$ &$\textbf{16.8}\pm 3.8$   \\
BeamRider & $684.8\pm 587.1$& $579.6 \pm 348.4$ & $846.1\pm 92.2 $ &$551.6\pm 25.1 $   \\
Enduro & $0\pm 0$& $8.1 \pm 8.4$ & $45.4 \pm 63.5$ &$9.7 \pm 6.9$   \\
SpaceInvaders & $35.8\pm 58.2$& $135 \pm 134.4$ & $\textbf{492.5} \pm 61.4$ &$490.4 \pm 32.3$   \\ 
Freeway & $0\pm 0$& $0 \pm 0$ & $\textbf{27.8} \pm 2.9$ &$17.1 \pm 12.9$ \\
Qbert & $1085.2\pm 0.5$&  $289.1
 \pm 0$ &$2030.2\pm 383.9$ &$\textbf{5659.2}\pm 1271.5$   \\
Seaquest & $955.2\pm 587.1$& $200.3 \pm 348.4$ & $718.1\pm 112.1 $ &$1297.3\pm 359.6 $   \\
Breakout & $\textbf{65.5}\pm 0$& $1 \pm 8.4$ & $39.7 \pm 4.2$ &$34.7 \pm 23.5$   \\
MsPacman & $643.2\pm 58.2$& $60.4 \pm 134.4$ & $1180.2 \pm 245.9$ &$1289.2 \pm 106.7$   \\ 
\midrule
\end{tabular}
\end{table*}

\begin{table*}[t!]
\centering
\scriptsize
\caption{\textbf{Distilled structured representations significantly enhance the performance upon convergence.} After training for three million steps, the performance of SA-Neural, SPACE-Neural, Coor-Neural, and Neural is evaluated. Coor-Neural consistently outperforms SA-Neural and SPACE-Neural upon convergence.}
\label{sa-3M}
\begin{tabular}{lcccc}
\toprule
Tasks & SA-Neural & SPACE-Neural  & Coor-Neural & Neural \\ \midrule
Pong & $17.2\pm 3$&  $-20.6
 \pm 0$ &$19.7\pm 0$ &$\textbf{19.9}\pm 0.3$   \\
Freeway & $19.8\pm 15$& $12.2 \pm 3$ & $\textbf{31.6} \pm 0.1$ &$25.3 \pm 6.8$ \\
\midrule
\end{tabular}
\end{table*}

\textbf{Distilled Structured Representations Significantly Enhance Performance}\quad These methods were trained for only one million steps due to the poor efficiency of SPACE-Neural. Similar experiment setup has been utilized by~\citet{10.1007/978-3-031-43421-1_36}.   We also report results for three million steps in  \Cref{sa-3M}, by which point the performances on Pong and Freeway had stabilized.
Note that compared to slot attention or the SPACE model, our perception module leads to better performance.
This is because our image-symbol dataset provides direct supervision for the location of objects. On the contrary, unsupervised methods are trained to reconstruct whole images and thus less focused on individual objects.

\subsection{Analysis of Coordinate Accuracy}\label{Analysis of Coordinate Accuracy}

\textbf{End-to-End Fine-Tuning Markedly Enhances Coordinate Prediction Accuracy in Policy-Relevant Tasks\quad}\Cref{cors} presents a quantitative assessment of the coordinate prediction accuracy achieved by our method. In \cref{cors-remain}, we document the experimental results for the additional four tasks. Notably, the F-MAE on Qbert, Seaquest, and MsPacman post-training markedly surpasses the pre-training figures. This suggests that end-to-end fine-tuning significantly improves accuracy in policy-relevant object coordinates.

\begin{table*}[t!]
\centering
\scriptsize
\caption{\textbf{INSIGHT demonstrates robustness to hyper-parameters.} MAE and F-MAE are used to measure coordinate prediction accuracy for Qbert, Seaquest, Breakout, and MsPacman, with all numbers scaled up by a factor of 100.}
\label{cors-remain}
\begin{tabular}{lcccccccccc}
\toprule
 &\multicolumn{2}{c}{INSIGHT}             &\multicolumn{2}{c}{w/o Pretrain}  &\multicolumn{2}{c}{Fix}           &\multicolumn{2}{c}{w/o NG}     &\multicolumn{2}{c}{Coor-Neural}   \\ 
Tasks &MAE &F-MAE &MAE &F-MAE &MAE &F-MAE &MAE &F-MAE &MAE &F-MAE \\ \midrule
Qbert            &$1.3\pm0$ &$\textbf{0.8}\pm0$       &$1.4\pm0.1$ &$0.9\pm0.1$      &$\textbf{1}\pm0$ &$0.9\pm0.1$  &$1.4\pm0.1$  &$1.4\pm0.1$  &$1.7\pm 0.1$ &/     \\
Seaquest       &$4.2\pm0.1$ &$\textbf{1.9}\pm0.2$    &$4.5\pm0.1$ &$2\pm0.2$      &$\textbf{4.1}\pm0.1$ &$3.7\pm0.3$     &$4.4\pm0.1$&$4.4\pm0.1$  &$4.5\pm 0.1$  &/     \\
Breakout          &$7.3\pm0.4$ &$5.4\pm0.6$     &$6.4\pm0.1$&$4.2\pm0.3$       &$\textbf{4.5}\pm0.2$   &$\textbf{2.5}\pm0.1$    &$5.2\pm0.3$ &$5.2\pm0.3$  &$6.7\pm 0.7$    &/       \\
MsPacman   &$7.2\pm0.1$ &$\textbf{4.3}\pm0.5$      &$6.9\pm0.2$&$5.1\pm0.4$       &$\textbf{6.5}\pm0.2$   &$6.8\pm0.7$    &$7.2\pm0.3$ &$7.2\pm0.3$  &$6.9\pm 0.3$    &/        \\ 
\midrule
\end{tabular}
\end{table*}

\begin{figure}[t!]
  \centering
  \includegraphics[width=0.4\linewidth]{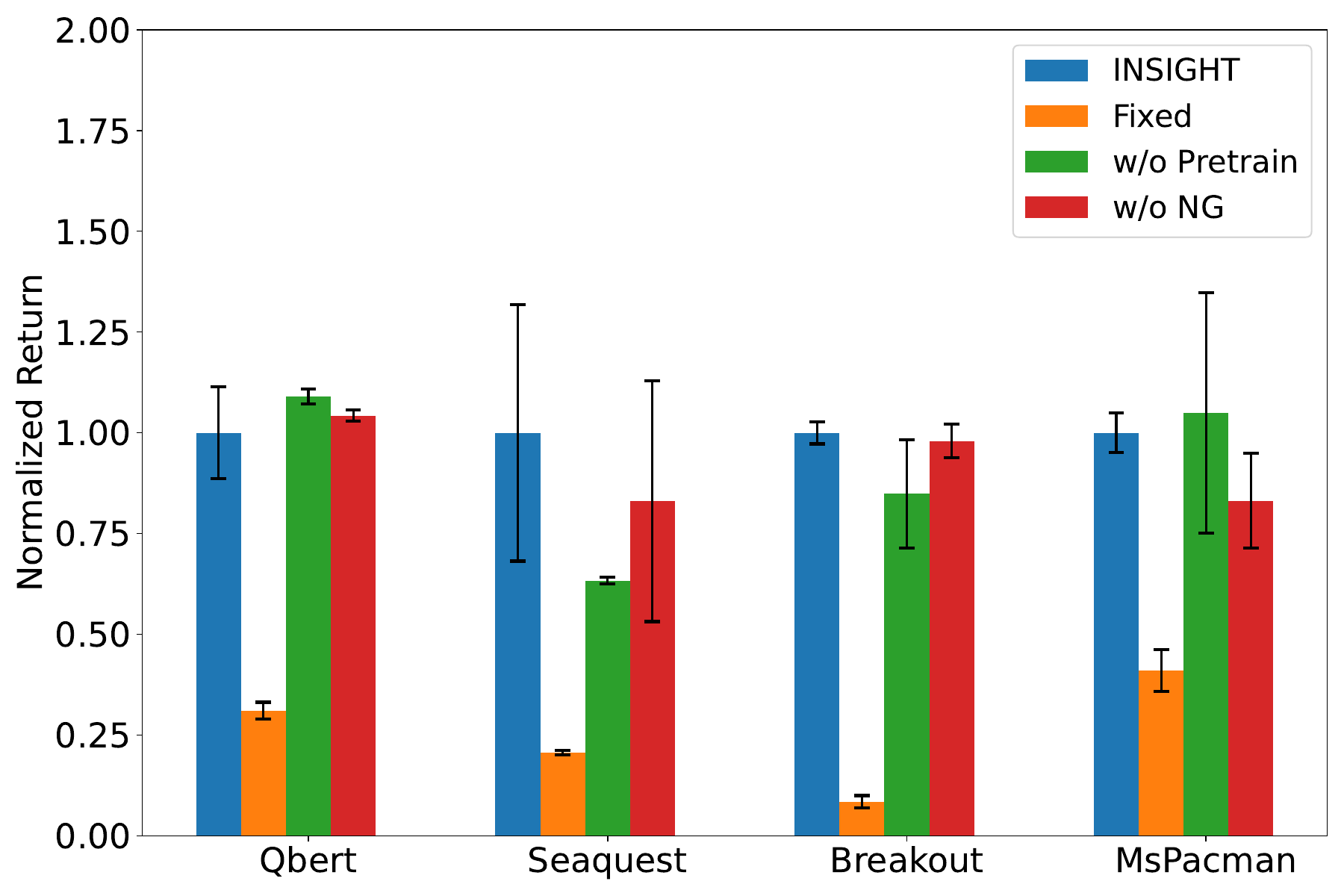}
\caption{\textbf{Each component of INSIGHT is critical for overall performance.} Detailed performance analyses of \ac{INSIGHT} and its variants across four tasks are presented. The findings in these four environments align with those in the other five environments.}
\label{remain_methodablation}
\end{figure}

\begin{figure}[t!]
  \centering
  \begin{subfigure}{0.33\linewidth}
    \includegraphics[width=\linewidth]{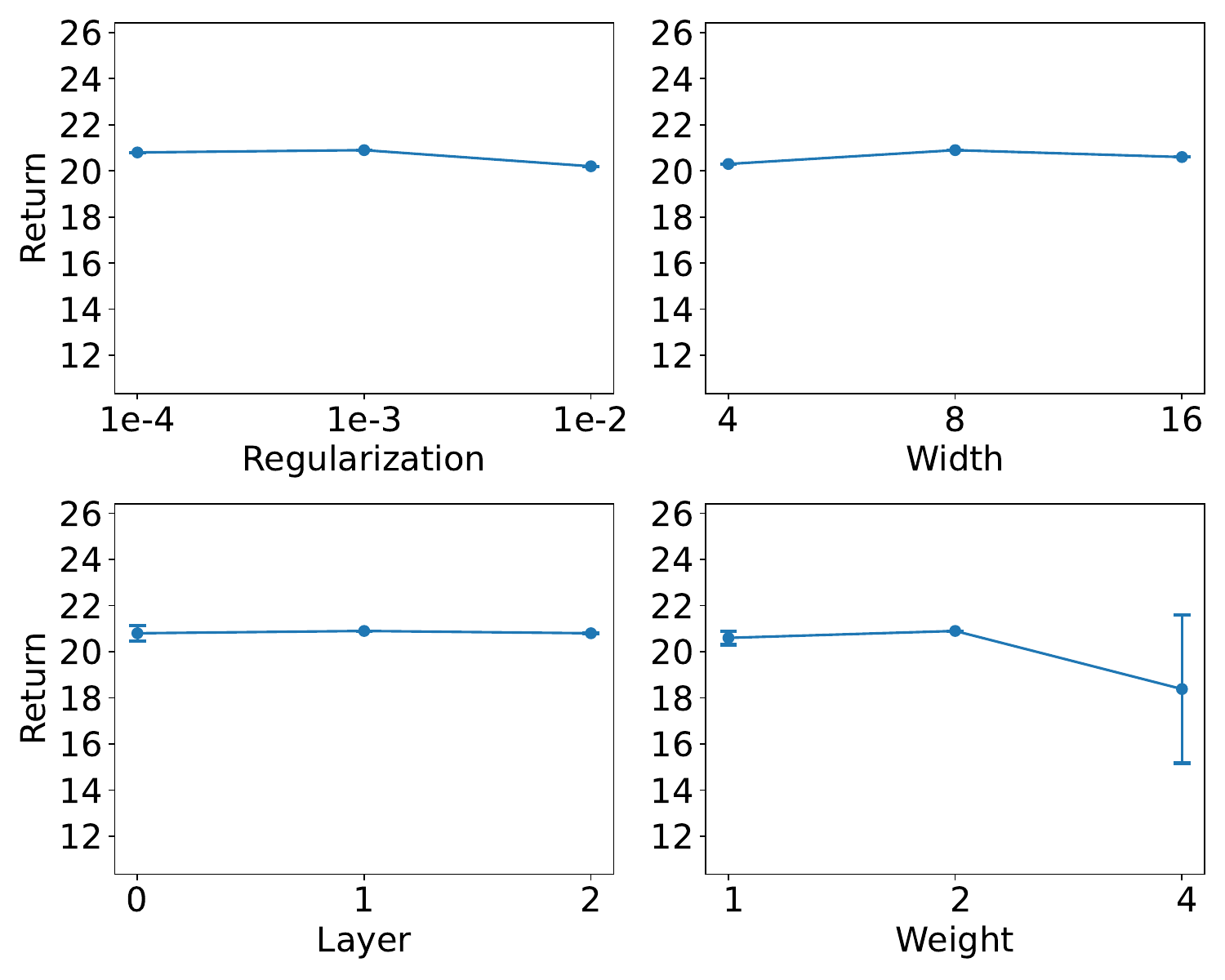}
    \caption{Pong.}
    \label{Pong-hyperablation}
  \end{subfigure}
  \hfill
  \begin{subfigure}{0.33\linewidth}
    \includegraphics[width=\linewidth]{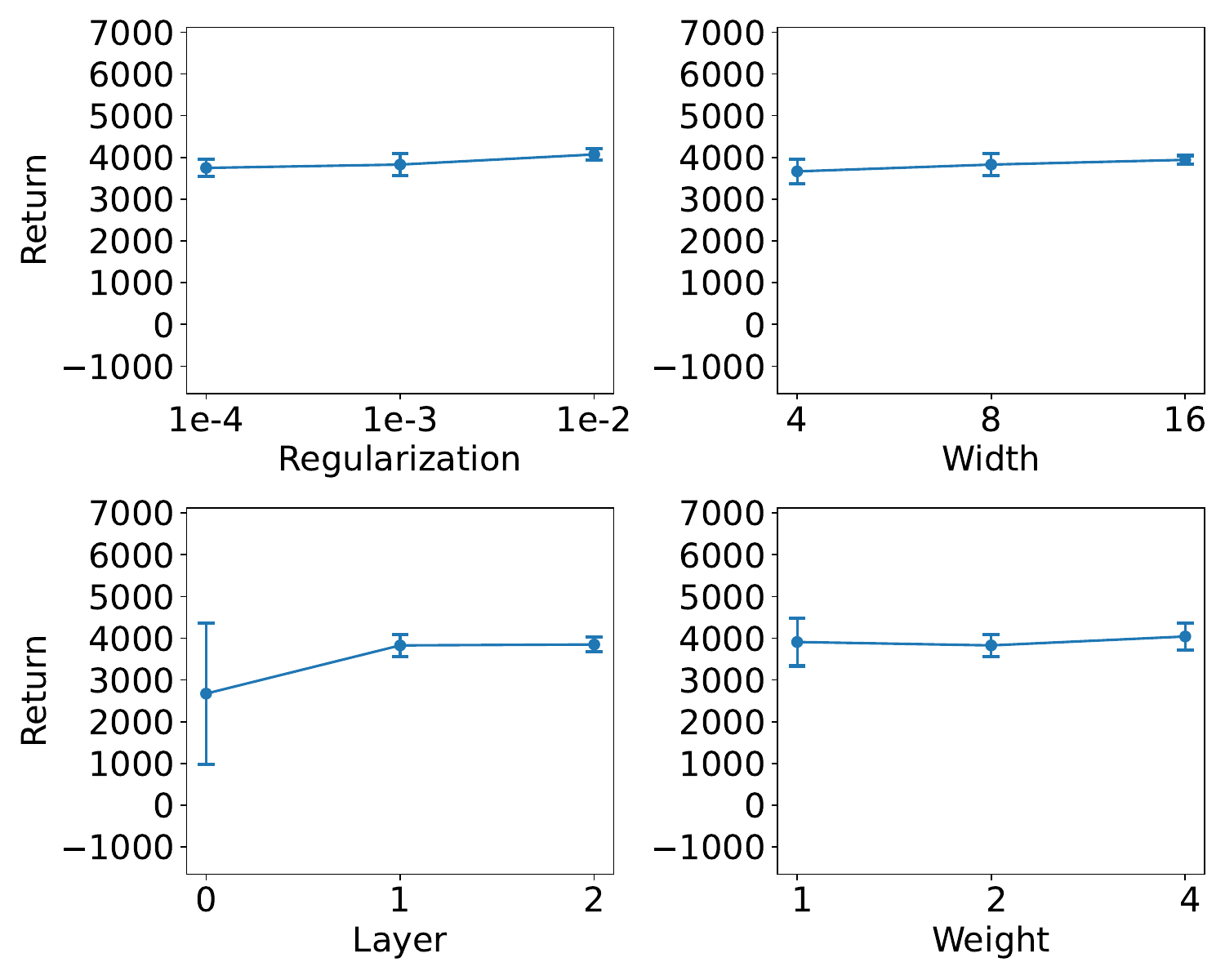}
    \caption{BeamRider.}
    \label{BeamRider-hyperablation}
  \end{subfigure}
  \begin{subfigure}{0.33\linewidth}
    \includegraphics[width=\linewidth]{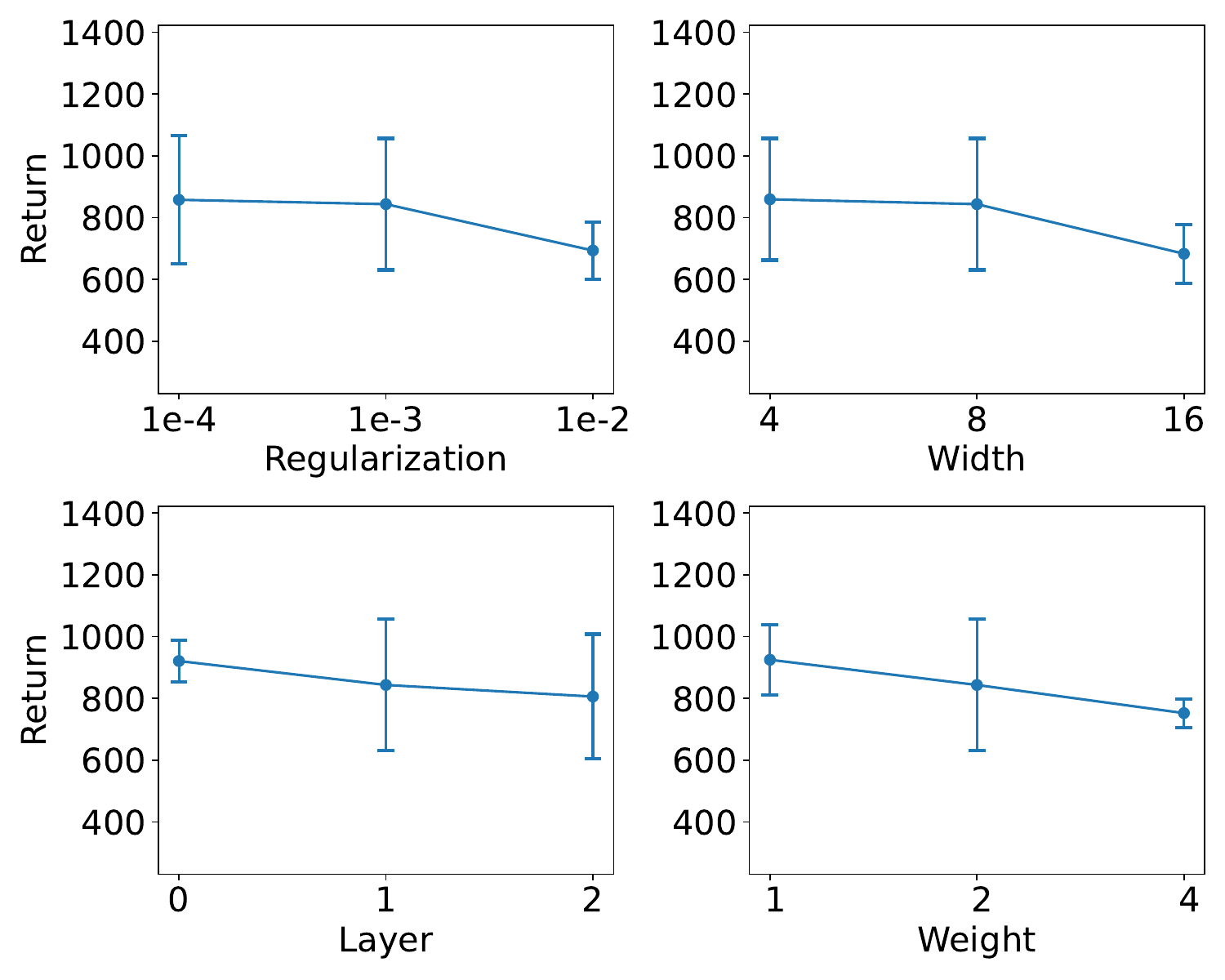}
    \caption{Enduro.}
    \label{Enduro-hyperablation}
  \end{subfigure}
  \\
  \begin{subfigure}{0.33\linewidth}
    \includegraphics[width=\linewidth]{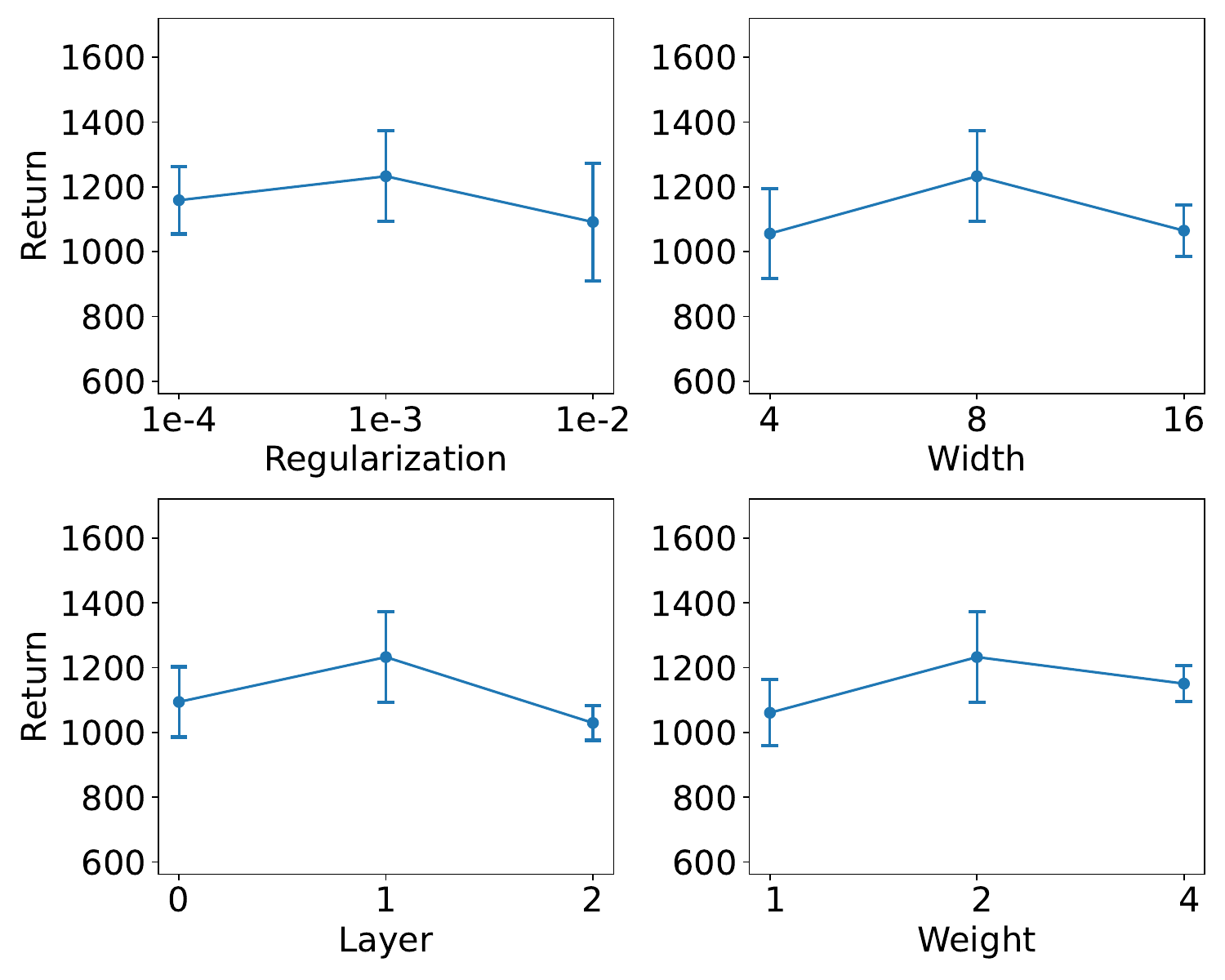}
    \caption{SpaceInvaders.}
    \label{SpaceInvaders-hyperablation}
  \end{subfigure}
  \hfill
  \begin{subfigure}{0.33\linewidth}
    \includegraphics[width=\linewidth]{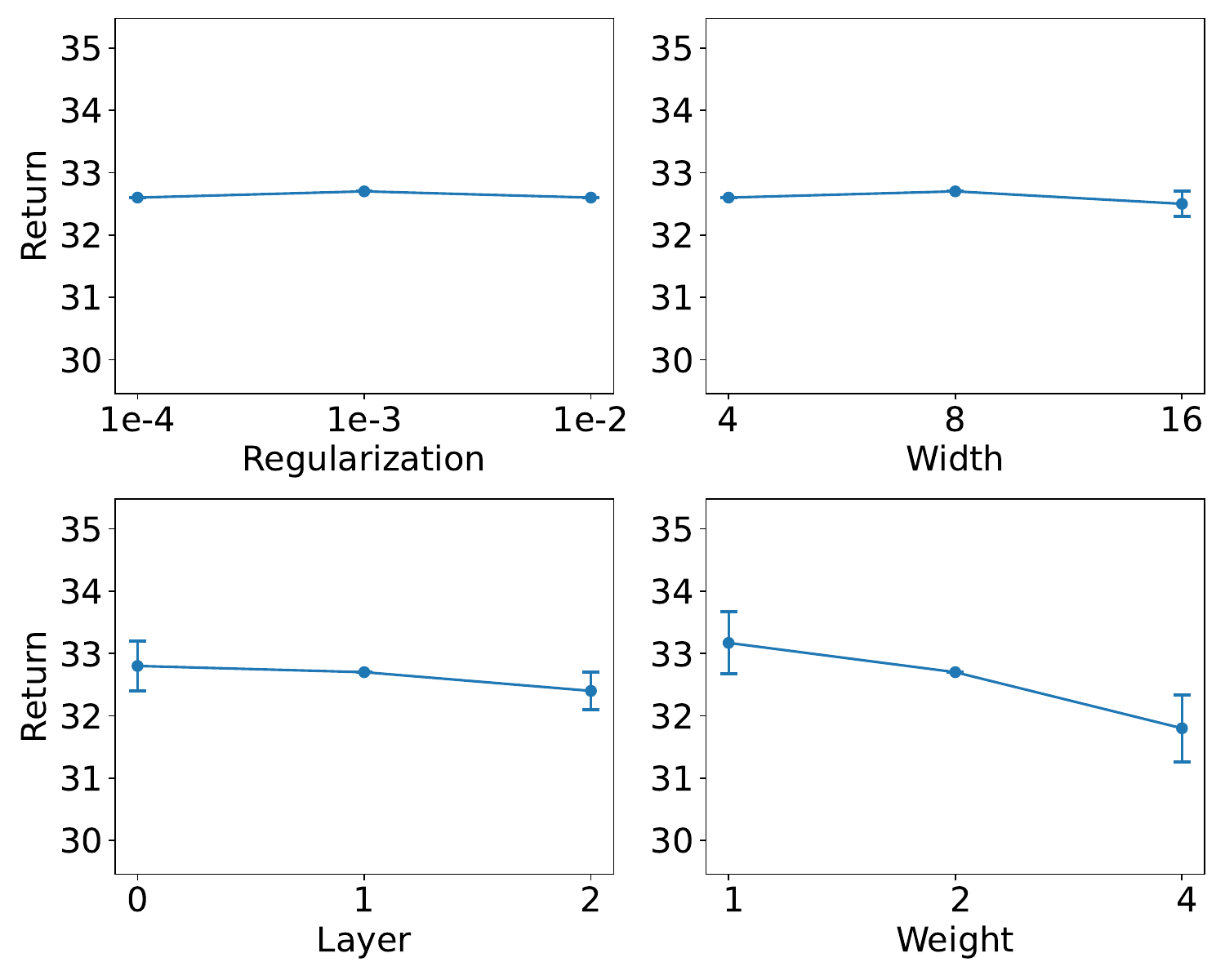}
    \caption{FreeWay.}
    \label{FreeWay-hyperablation}
  \end{subfigure}
  \begin{subfigure}{0.33\linewidth}
    \includegraphics[width=\linewidth]{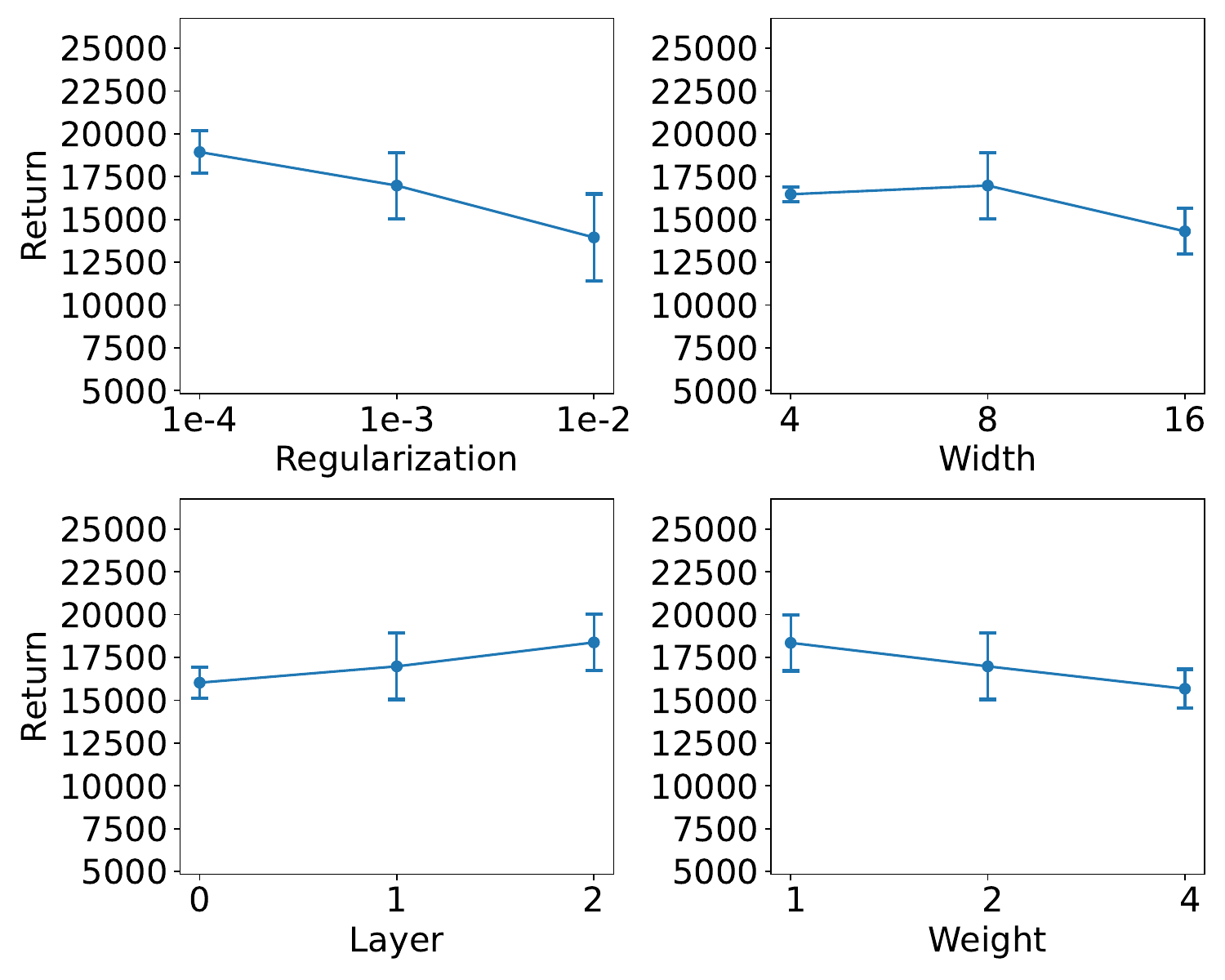}
    \caption{Qbert.}
    \label{Qbert-hyperablation}
  \end{subfigure}
  \hfill
  \begin{subfigure}{0.33\linewidth}
    \includegraphics[width=\linewidth]{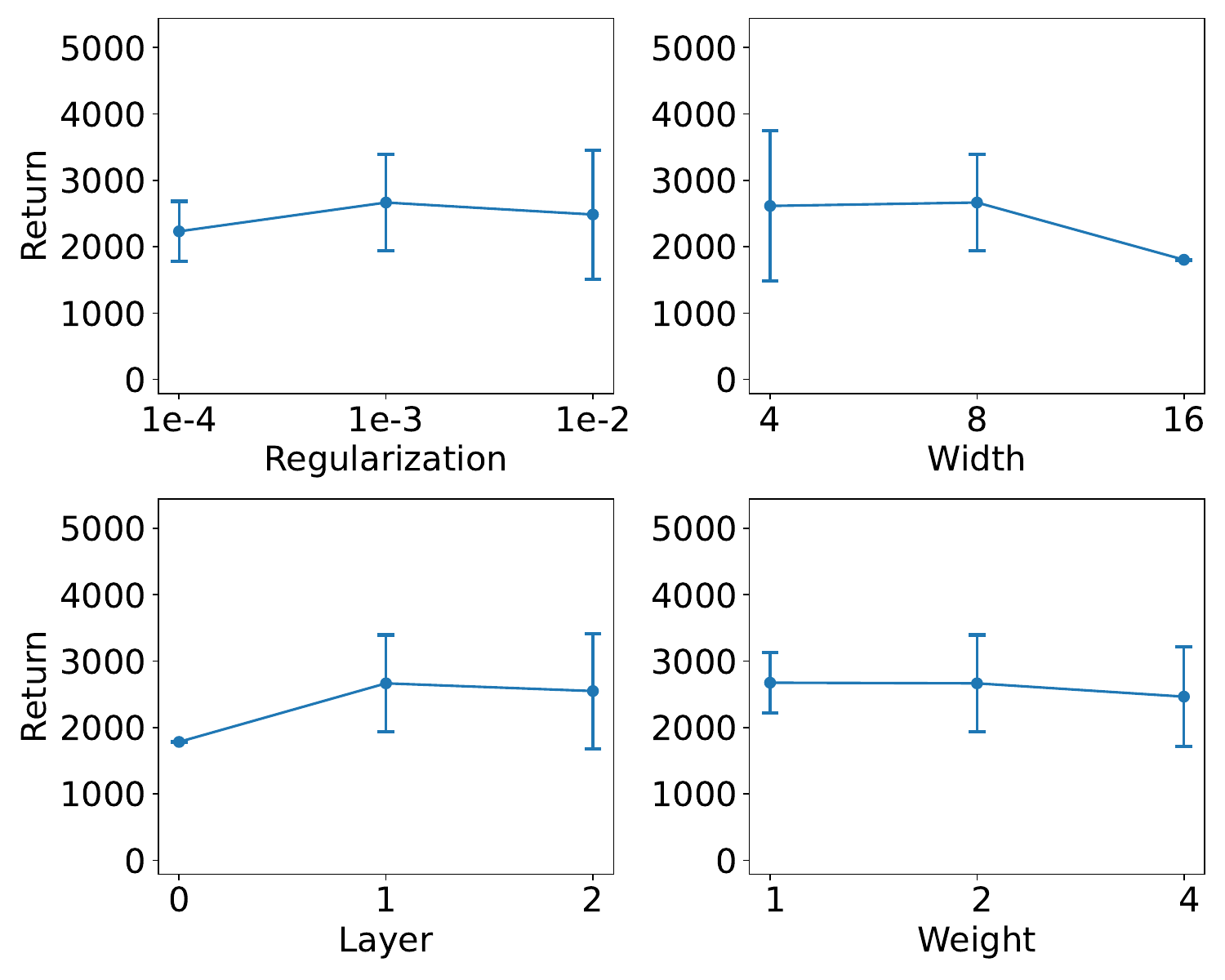}
    \caption{Seaquest.}
    \label{Seaquest-hyperablation}
  \end{subfigure}
  \hfill
  \begin{subfigure}{0.33\linewidth}
    \includegraphics[width=\linewidth]{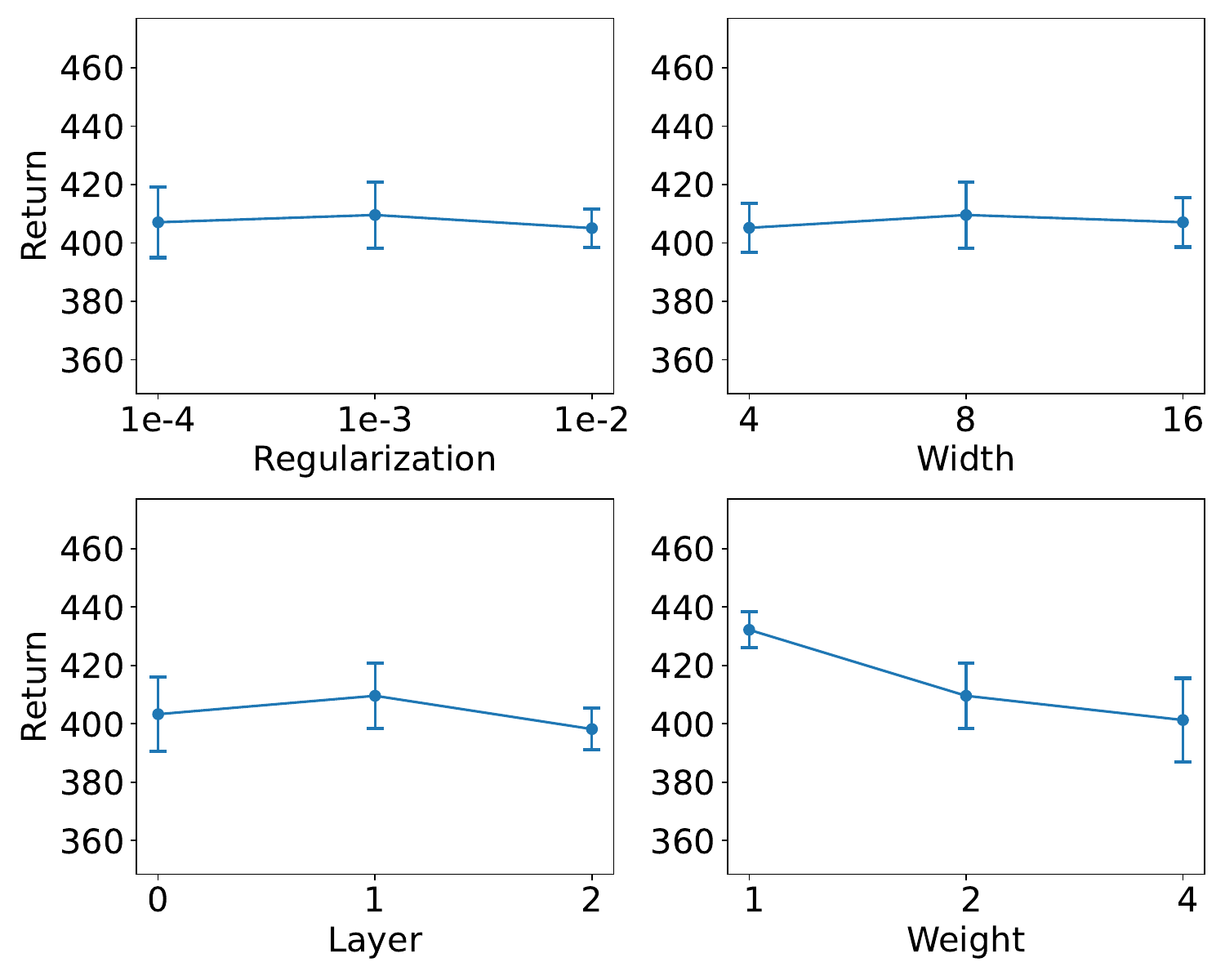}
    \caption{Breakout.}
    \label{Breakout-hyperablation}
  \end{subfigure}
  \hfill
  \begin{subfigure}{0.33\linewidth}
    \includegraphics[width=\linewidth]{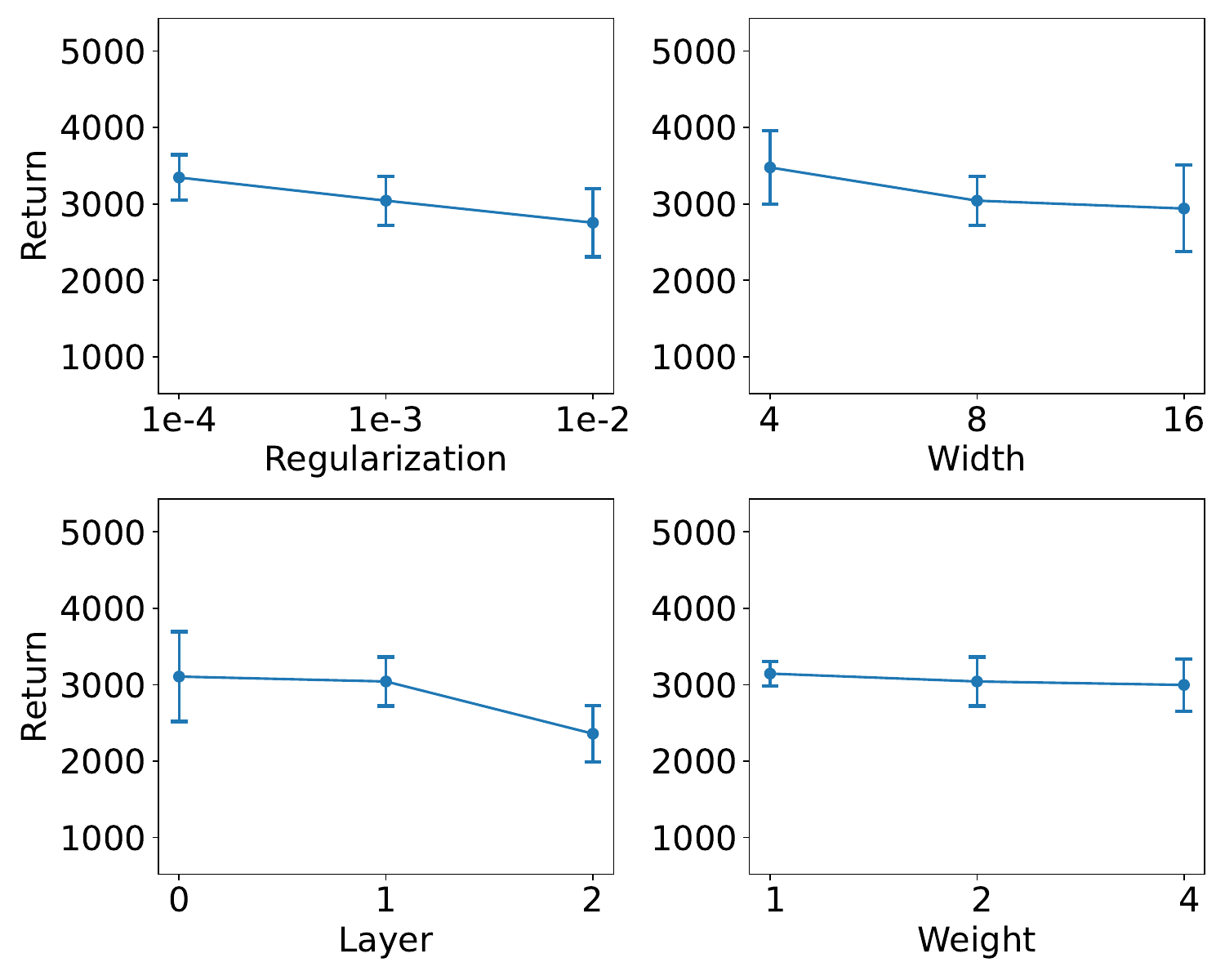}
    \caption{MsPacman.}
    \label{MsPacman-hyperablation}
  \end{subfigure}
\caption{\textbf{INSIGHT demonstrates robustness to hyper-parameters.} Examining the effects of the regularization coefficient $\lambda_\text{reg}$, the width and number of hidden layers of the EQL actor, and the weight of $\mathcal{L}_\text{CNN}$ on performance. INSIGHT shows substantial robustness to variations in these hyper-parameters.}
  \label{hyper ablation remain}
\end{figure}

\subsection{Extended Ablation Study Details}\label{Expanded Ablation Study Details}

\textbf{
Generalizability of Ablation Experiment Conclusions Across Tasks\quad}This subsection extends the ablation studies detailed in \cref{methodablation} and \cref{hyperablation} to encompass all tasks, as referenced in \cref{remain_methodablation}-\cref{MsPacman-hyperablation}. Notably, the comprehensive ablation analysis presented in \cref{remain_methodablation} demonstrates performance declines in most tasks when methods are modified. Furthermore, the consistency observed across various hyperparameters, as shown in \cref{Pong-hyperablation}-\cref{MsPacman-hyperablation}, corroborates the findings discussed in \cref{subsec:task_performance}.

\begin{table*}[t!]
\centering
\scriptsize
\caption{\textbf{INSIGHT outperforms CleanRL.} Evaluating the performance of the proposed INSIGHT, Neural, and CleanRL (Huang et al., 2022) across nine Atari tasks. INSIGHT matches Neural's performance on all tasks and surpasses all NS-RL baselines. In most instances, our neural baseline competes closely with the CleanRL baseline.}
\label{Performance-cleanrl}
\begin{threeparttable}
\begin{tabular}{lccc}
\toprule
Task & \ac{INSIGHT} & Neural & CleanRL\\ \midrule
Pong & $\textbf{20.9}\pm0.1$&$20.4\pm0.6$& $20.7\pm0.4$    \\
BeamRider &  $3828.1\pm261.1$ &$\textbf{3868.1}\pm204.2$ & $2270.5\pm107.3$  \\
Enduro & $843.7\pm213.8$ &$676.9\pm730.4$ & $\textbf{1006.6}\pm244.1$  \\

Qbert &  
$16978.6\pm1936.1$   & $\textbf{17879.2}\pm1857.1$ & $16619.8\pm256.3$ \\

SpaceInvaders & $\textbf{1232.6}\pm140.7$ &$1184.6\pm137.6$ & $1023.1\pm160.2$  \\

Seaquest & $\textbf{2665.7}\pm728.2$ &$1804.8\pm20.1$ & $1841.2\pm17.1$ \\
Breakout & $\textbf{409.6}\pm11.3$ &$259.6\pm183.8$ & $387.8\pm11.8$ \\

Freeway & $\textbf{32.7} \pm 0.1$ &$28.7\pm5.3$ & $32.2\pm1.1$  \\
MsPacman & $\textbf{3042.5}\pm320.1$ &$2737.1\pm562.3$ & $2237.2\pm163.5$ \\
\bottomrule
\end{tabular}
\end{threeparttable}
\end{table*}

\begin{figure}[t!]
  \centering
  \begin{subfigure}{0.245\linewidth}
    \includegraphics[width=\linewidth]{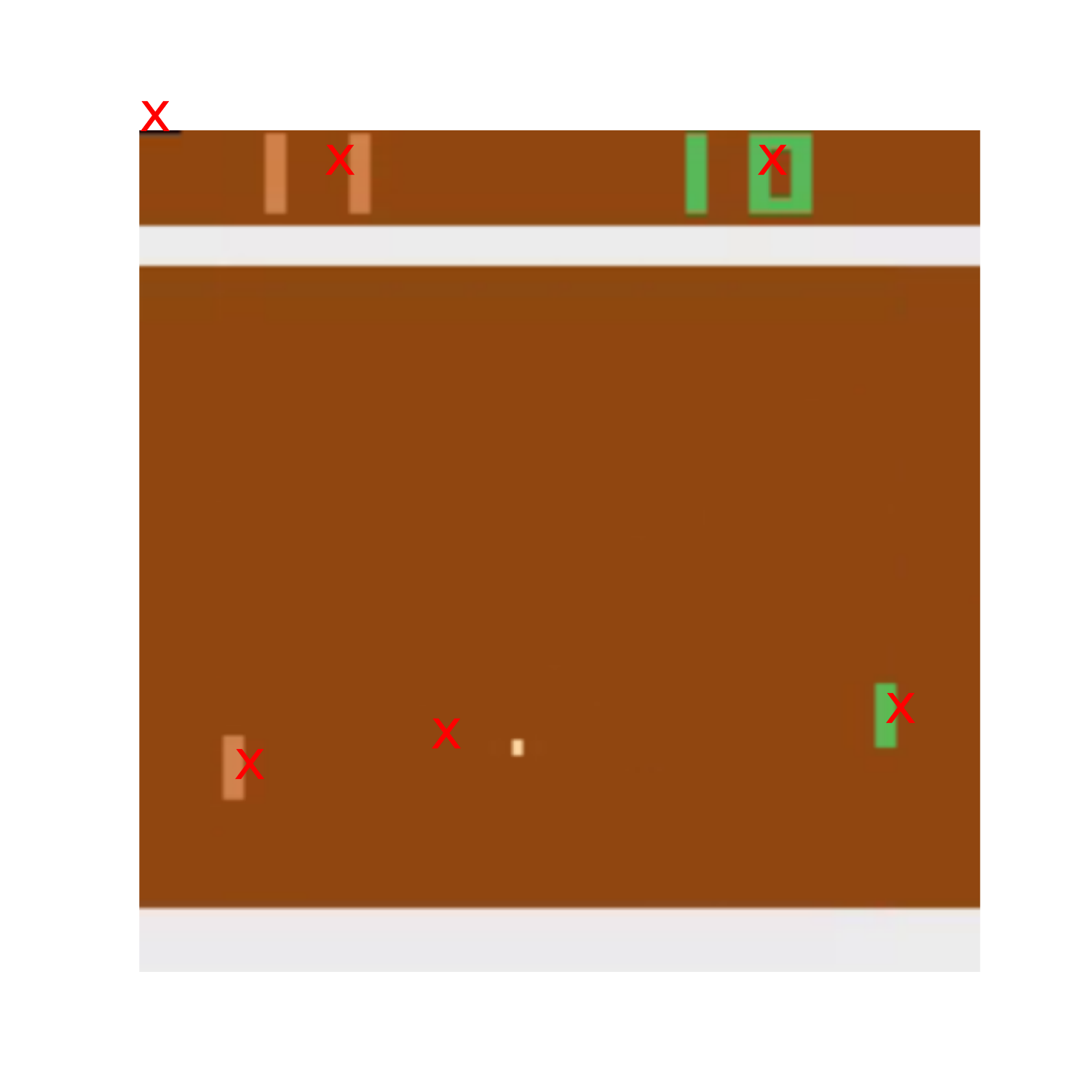}
    \caption{Pong (before).}
    \label{Pong-coor-before}
  \end{subfigure}
  \begin{subfigure}{0.245\linewidth}
    \includegraphics[width=\linewidth]{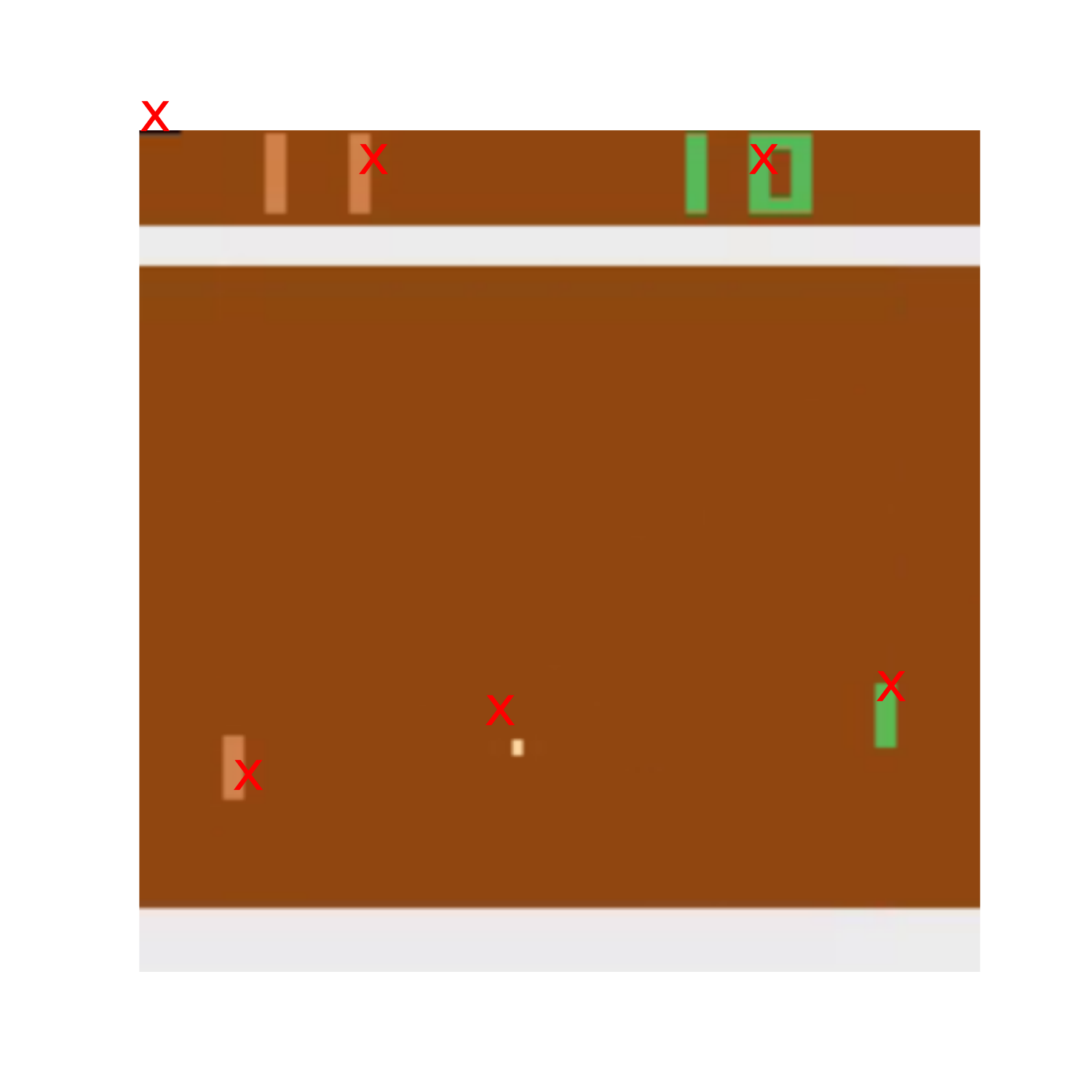}
    \caption{Pong (after).}
    \label{Pong-coor}
  \end{subfigure}
  \begin{subfigure}{0.245\linewidth}
    \includegraphics[width=\linewidth]{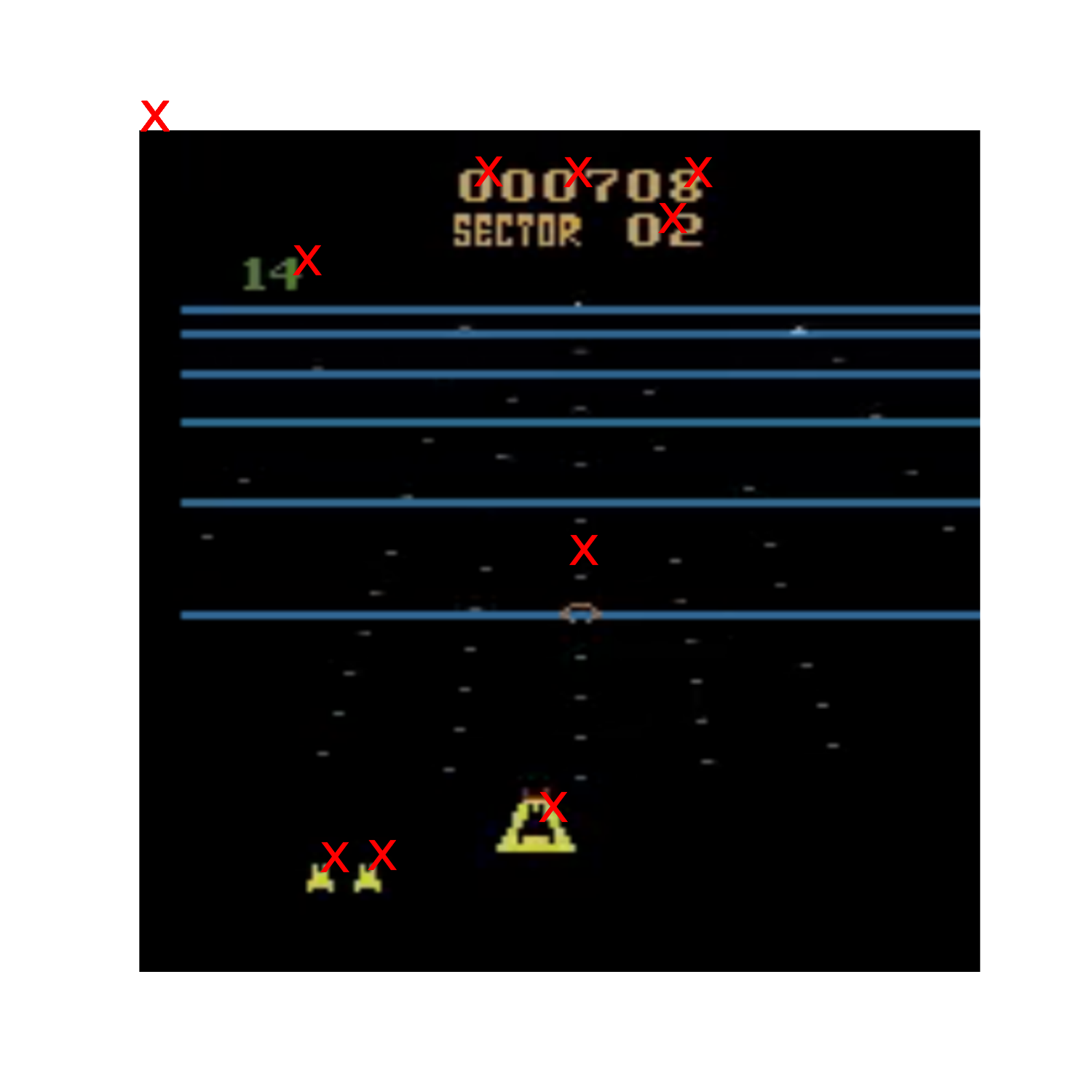}
    \caption{BeamRider (before).}
    \label{BeamRider-coor-before}
  \end{subfigure}
  \begin{subfigure}{0.245\linewidth}
    \includegraphics[width=\linewidth]{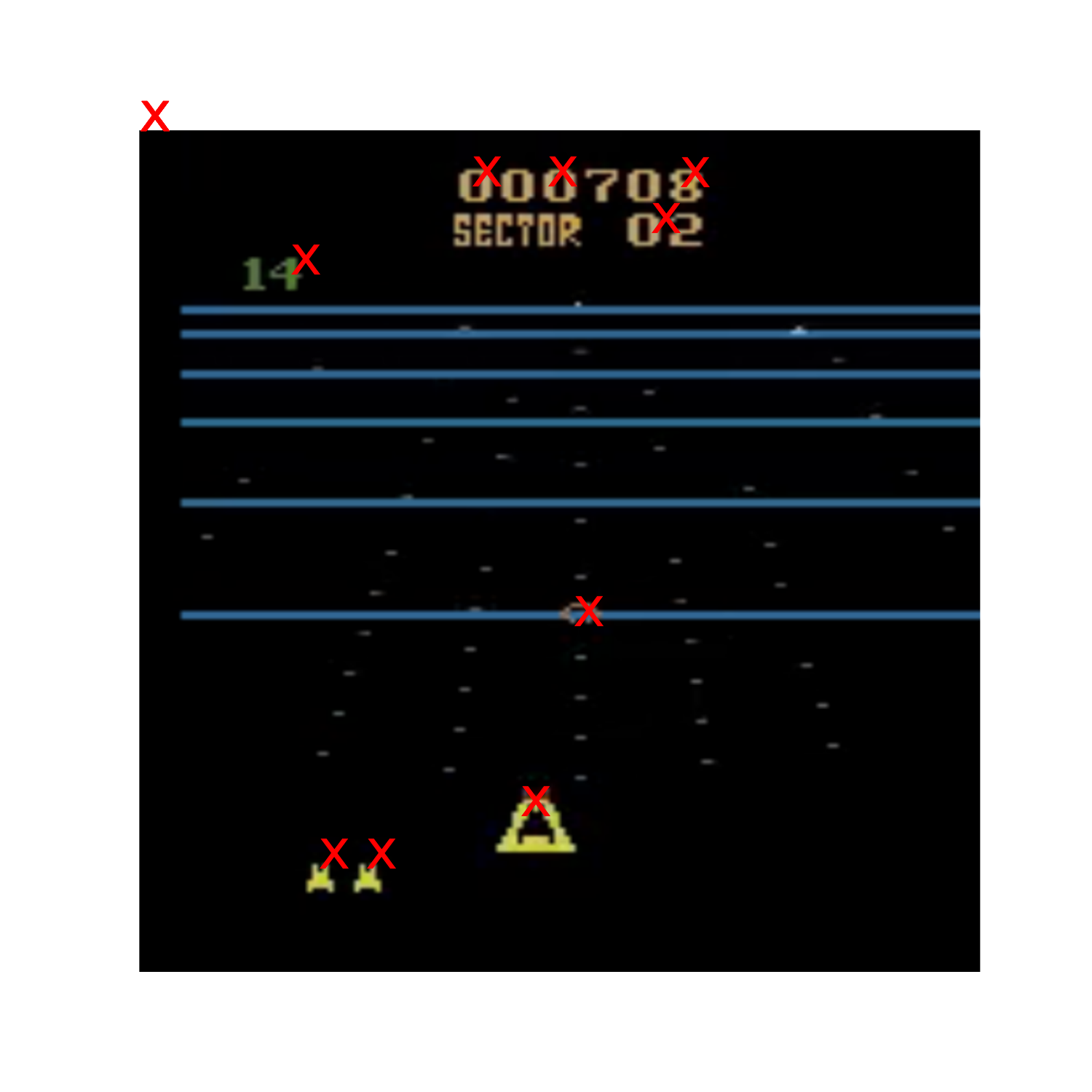}
    \caption{BeamRider (after).}
    \label{BeamRider-coor}
  \end{subfigure}
\\
  \begin{subfigure}{0.245\linewidth}
    \includegraphics[width=\linewidth]{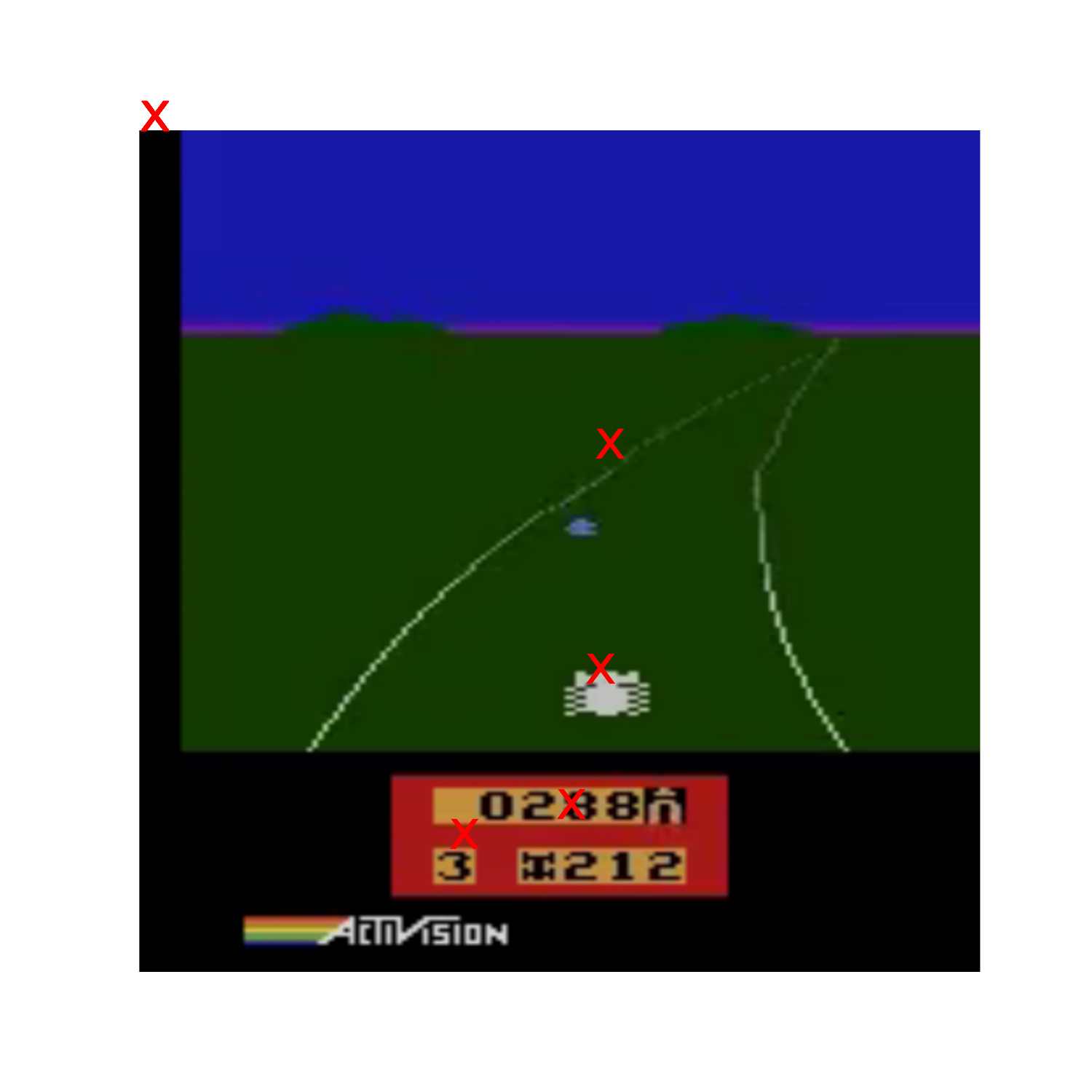}
    \caption{Enduro (before).}
    \label{Enduro-coor-before}
  \end{subfigure}
  \begin{subfigure}{0.245\linewidth}
    \includegraphics[width=\linewidth]{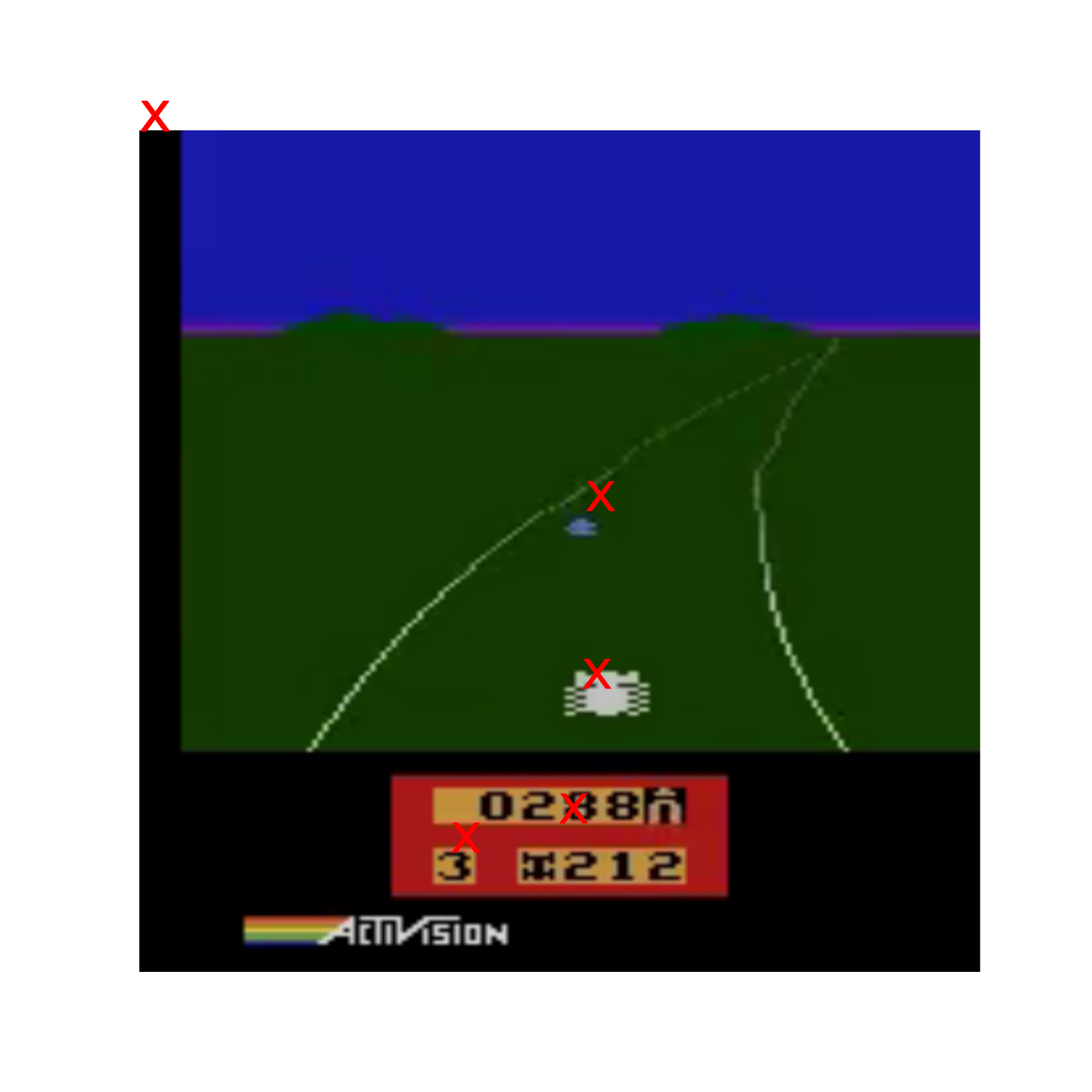}
    \caption{Enduro (after).}
    \label{Enduro-coor}
  \end{subfigure}
  \begin{subfigure}{0.245\linewidth}
    \includegraphics[width=\linewidth]{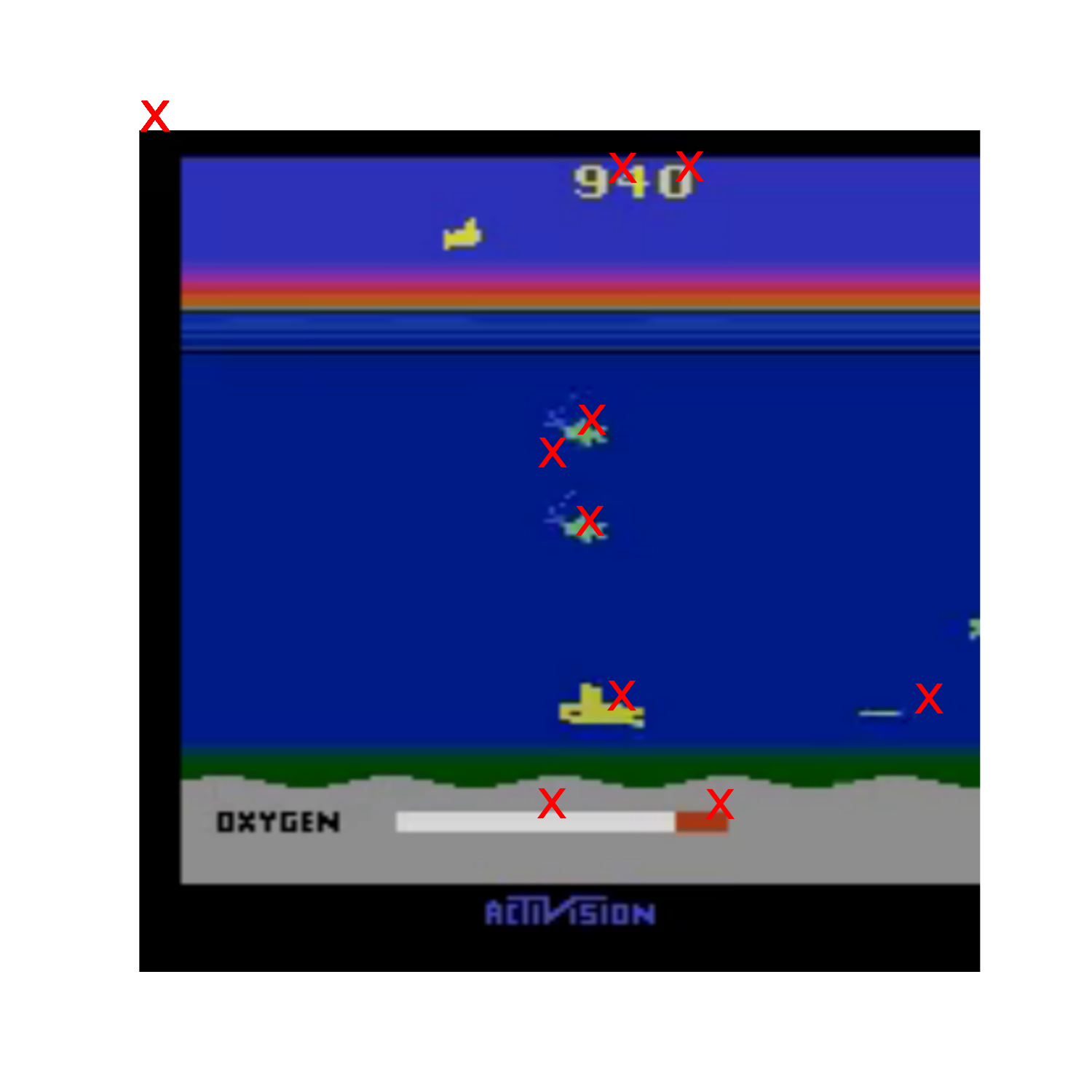}
    \caption{Seaquest (before).}
    \label{Seaquest-coor-before}
  \end{subfigure}
  \begin{subfigure}{0.245\linewidth}
    \includegraphics[width=\linewidth]{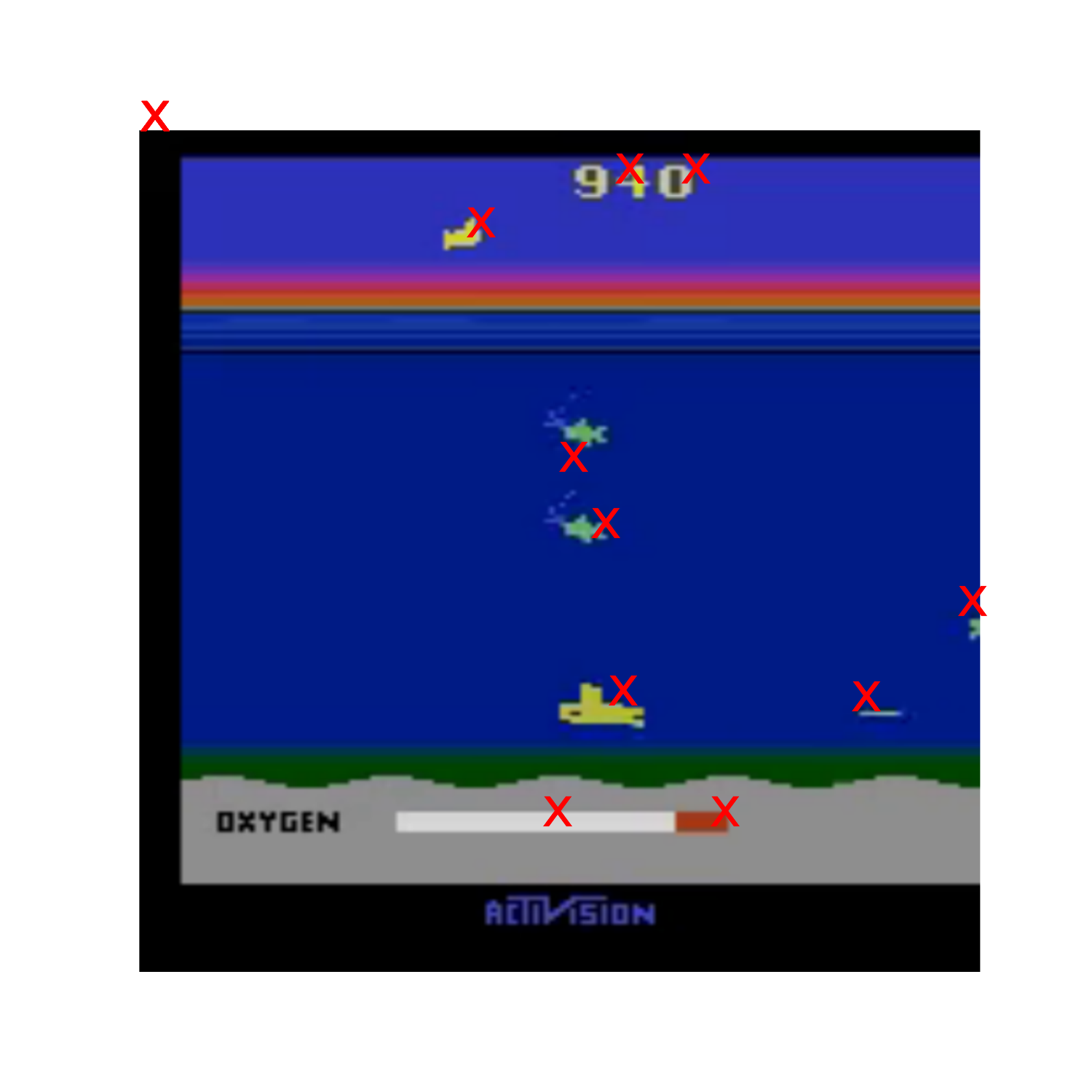}
    \caption{Seaquest (after).}
    \label{Seaquest-coor}
  \end{subfigure}
  \\
  
  \begin{subfigure}{0.245\linewidth}
    \includegraphics[width=\linewidth]{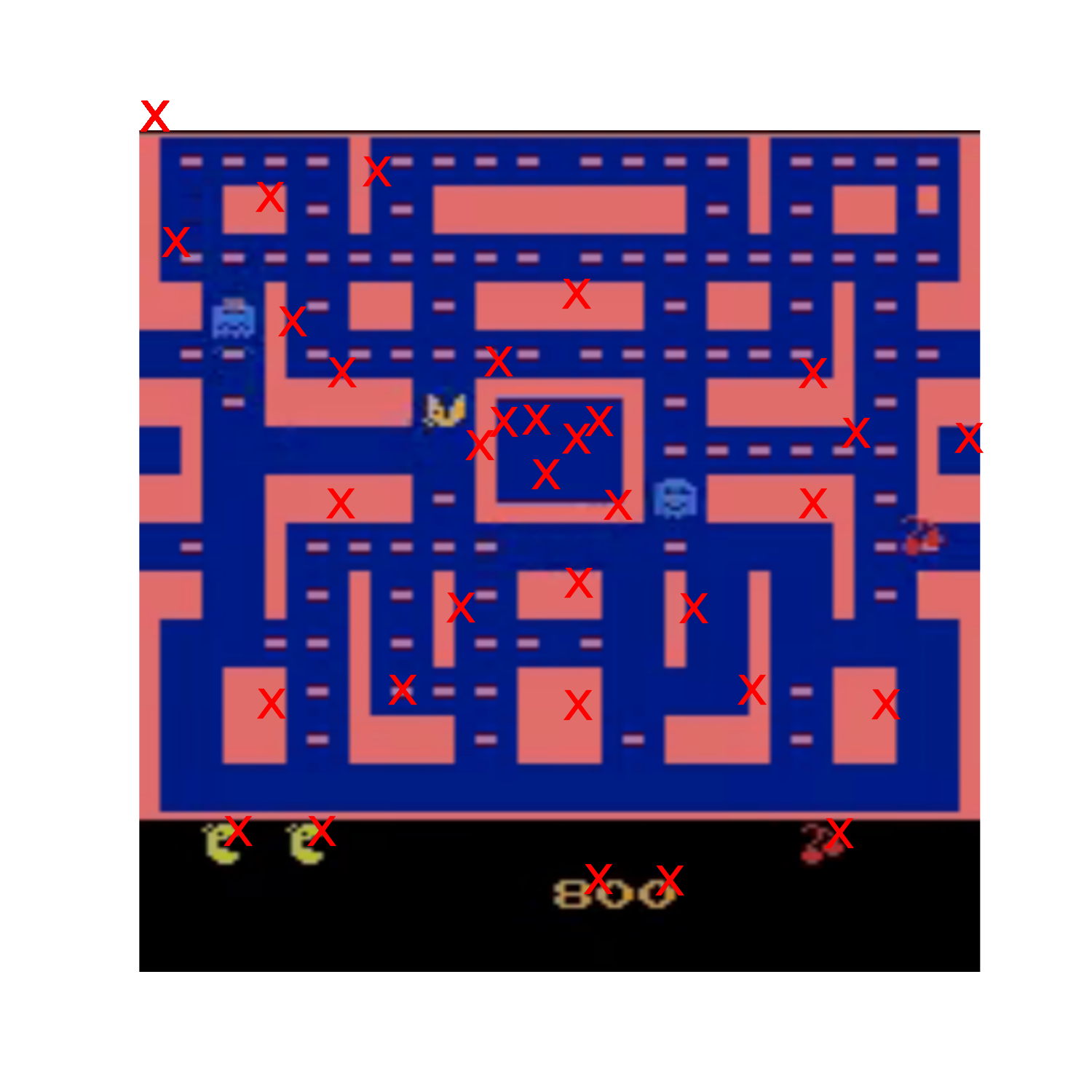}
    \caption{MsPacman (before).}
    \label{MsPacman-coor-before}
  \end{subfigure}
  \begin{subfigure}{0.245\linewidth}
    \includegraphics[width=\linewidth]{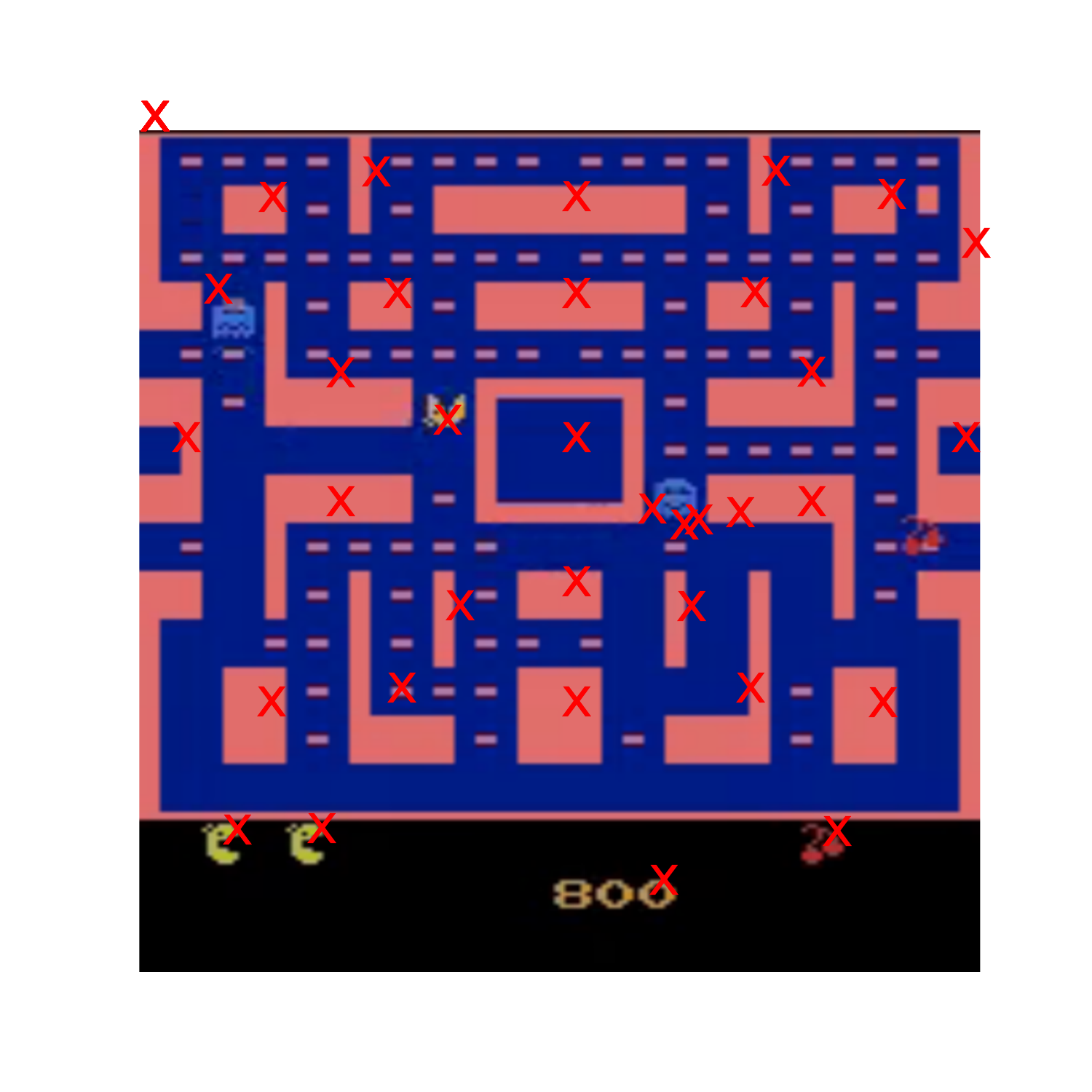}
    \caption{MsPacman (after).}
    \label{MsPacman-coor}
  \end{subfigure}
    \hfill
    \caption{Predicted coordinates before and after policy learning.}
    \label{Coordinates before and after training}
\end{figure}

\subsection{Additional Baseline}\label{Additional Baseline}

\textbf{
INSIGHT Outperforms CleanRL\quad}In this section, we introduce an additional baseline, CleanRL, to evaluate the performance of INSIGHT. CleanRL is an open-source implementation derived from the CleanRL library \citep{huang2022cleanrl}. The sole modification involves adjusting the CNN's output dimension from 512 to 2048 or higher to facilitate a fair comparison with our framework. It is noteworthy that our neural baseline exhibits competitive performance relative to CleanRL. As shown in \cref{Performance-cleanrl}, INSIGHT achieves superior returns in six tasks, unequivocally demonstrating its capability to outperform existing neural policies in online environments.

\subsection{Visualized Coordinate Tracking}\label{Visualized coordinate tracking}

In this section, we expand upon the advantages of end-to-end training for coordinate prediction, initially presented in \cref{cors}, by offering a more intuitive visualization. \cref{Coordinates before and after training} illustrates the object coordinates before and after the training process. For instance, in \cref{Pong-coor,Pong-coor-before}, we note a marked reduction in the pixel coordinate shift of the ball post-training. Similarly, the laser emitted by the agent in \cref{BeamRider-coor,BeamRider-coor-before} and the car on the road in \cref{Enduro-coor,Enduro-coor-before} exhibit significantly diminished pixel coordinate shifts following training. Moreover, post-training observations in \cref{Seaquest-coor,Seaquest-coor-before} reveal the successful detection of enemy coordinates at the periphery, a detail that was previously unattainable. Lastly, in \cref{MsPacman-coor,MsPacman-coor-before}, the model demonstrates its ability to consistently track the blue enemy post-training, overcoming the initial limitation of losing track of the object.

\subsection{Accuracy Evaluation of the FastSAM and DeAot Method}\label{Evaluation of Accuracy on Frame-Symbol Datasets}

\textbf{Our Approach Competes with MOC Solely Through Zero-Shot Generalization\quad}To evaluate the reliability of our dataset generation method, we further evaluate the accuracy of bounding boxes in the OCAtari dataset ~\citep{delfosse2023ocatari}. The table below shows the F-score from the Pong, SpaceInvaders and MsPacman environments within the OCAtari dataset, compared with the results of SPACE+MOC, as reported in ~\citep{delfosse2023boosting}:

\begin{table}[ht]
\centering
\caption{\textbf{Our method competes with MOC solely through zero-shot generalization. } The F-score evaluation of the FastSAM and DeAot method on OCAtari.}
\begin{tabular}{lccc}
\toprule
  Env& INSIGHT & SPACE+MOC(w/o OC) & SPACE+MOC \\
\midrule
Pong & 97.5 &  91.5 & 87.4  \\
SpaceInvaders & 86.5 & 85.1 & 85.2  \\
MsPacman & 46.5 & 88.6 & 90.5 \\
\bottomrule
\end{tabular}
\end{table}

\begin{table}[ht]
\centering
\caption{\textbf{Our method achieves high recall solely through zero-shot generalization. } The precision and recall evaluation of the FastSAM and DeAot method on OCAtari.}
\label{table:ocatari_recall_evaluation}
\begin{tabular}{lcc}
\toprule
  Env& Precision & Recall  \\
\midrule
Pong & 96.3 & 98.7   \\
SpaceInvaders & 88.5 & 84.5   \\
MsPacman & 30.5 & 97.8  \\
\bottomrule
\end{tabular}
\end{table}

Note that INSIGHT achieves an impressive F-score of 97.5\% on Pong, significantly outperforming SPACE+MOC. For SpaceInvaders, we set IOU to 1 to segment as many different objects as possible. However, MsPacman’s F-score is lower than that achieved with the MOC-based method.

\textbf{Our Method Achieves High Recall Solely Through Zero-Shot Generalization\quad}To further analyze this, we provide the precision and recall scores in Table \ref{table:ocatari_recall_evaluation}. In the MsPacman environment, the very low precision and nearly perfect recall indicate that while all objects were segmented, many irrelevant objects not in the OCAtari dataset were also included. Given our goal during dataset generation is to segment as many complete objects as possible for downstream symbolic policy, higher precision may be more appropriate than F-score in RL scenarios.

\subsection{Additional Computation Time Information}\label{Additional Computation Time Information}

The average training time of each component on 9 Atari environments using NVidia RTX 3090 GPU is as follows:

\begin{table}[ht]
\centering
\caption{Training time of each component on 9 Atari environments.}
\begin{tabular}{lccc}
\toprule
 & Pretraining the Visual Perception & Policy learning & Policy Learning w/o Finetune \\
\midrule
Time (ms) &$2.6 \pm 0.3$ & $7.3 \pm 1.1$ & $6.2 \pm 0.8$ \\
\bottomrule
\end{tabular}
\end{table}

Note that the end-to-end training time of the visual module only accounts for about 1/10 of the total pipeline, but it greatly improves the return and prediction accuracy, which reflects the high profitability.

{\textbf{Overall Computation Time Increases with the Number of Objects}\quad} In \cref{table:seg_time}, we present the computation times of FastSAM, SAM, and DeAot with varying numbers of objects:
\begin{table}[ht!]
\centering
\caption{\textbf{Overall computation time increases with the number of objects. } The computation times (s) of FastSAM, SAM, and DeAot with varying numbers of objects.}
\label{table:seg_time}
\begin{tabular}{lccc}
\toprule
 Num & FastSAM & SAM & DeAot \\
\midrule
5 &$0.07 $ & $4.21$ & $0.01$ \\
10 &$0.07 $ & $4.57$ & $0.02$ \\
20 &$0.07 $ & $5.21$ & $0.02$ \\
50 &$0.08 $ & $5.83$ & $0.03$ \\
100 &$0.08 $ & $5.84$ & $0.07$ \\
\bottomrule
\end{tabular}
\end{table}
The segmentation speed of FastSAM remains relatively constant despite varying object counts, whereas DeAot's speed significantly decreases as more objects are introduced. This suggests an increase in overall computational overhead proportional to the number of objects.

The average inference time of each component on 9 Atari environments using NVidia RTX 3090 GPU is as follows

\begin{table}[ht]
\centering
\caption{Inference time of each component on 9 Atari environments.}
\begin{tabular}{lcc}
\toprule
 & CNN & EQL  \\
\midrule
Time (ms) &$0.3 $ & $ 1.7$  \\
\bottomrule
\end{tabular}
\end{table}

Note that we have adopted a simple CNN as the tracking module, which significantly enhances the inference speed of INSIGHT.

\subsection{Dataset segmentation results}\label{Data set segmentation results}
To visually demonstrate our segmentation accuracy, we took five frames from each Atari game, and summarized them in \cref{segmentation results}:

\begin{figure}[t!]
  \centering
  \begin{subfigure}{1\linewidth}
    \includegraphics[width=\linewidth]{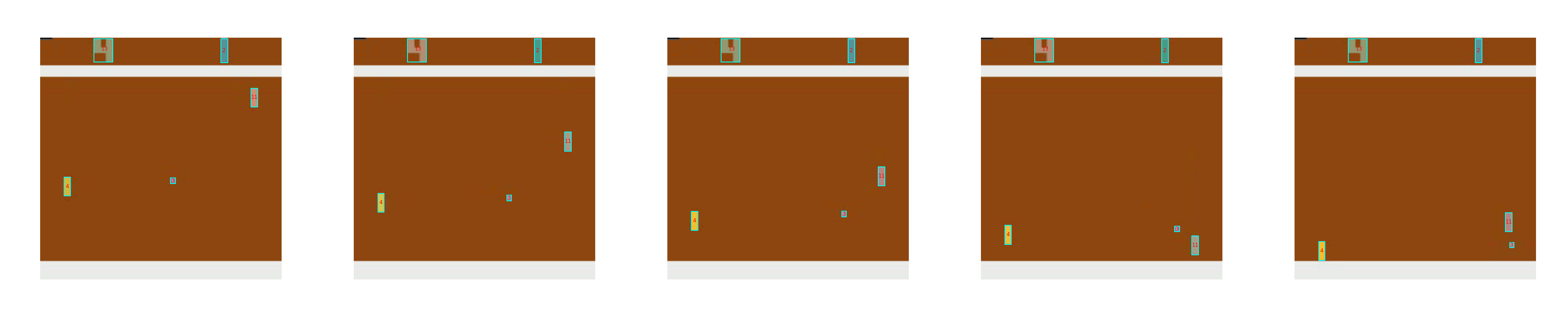}
    \caption{Pong.}
    \label{Pong_concatenated}
  \end{subfigure}
  \begin{subfigure}{1\linewidth}
    \includegraphics[width=\linewidth]{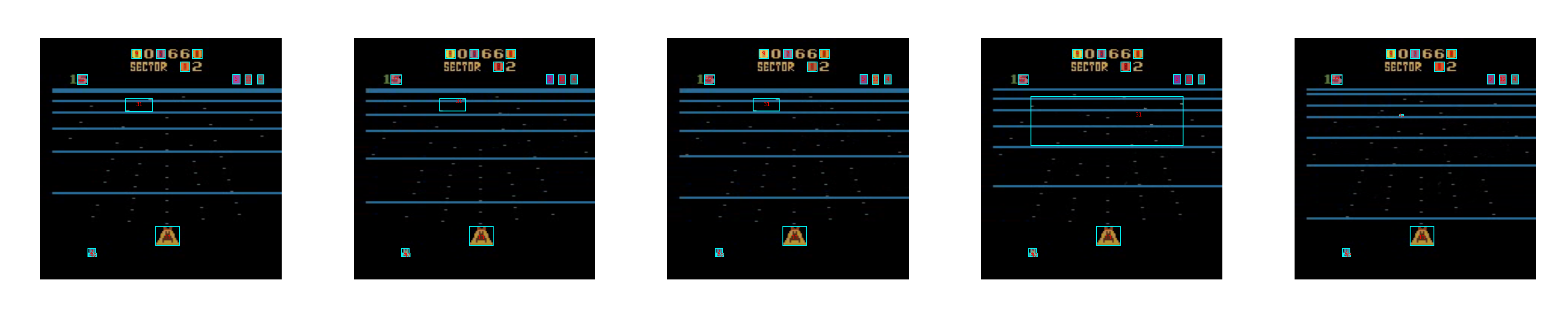}
    \caption{BeamRider.}
    \label{BeamRider_concatenated}
  \end{subfigure}
  \begin{subfigure}{1\linewidth}
    \includegraphics[width=\linewidth]{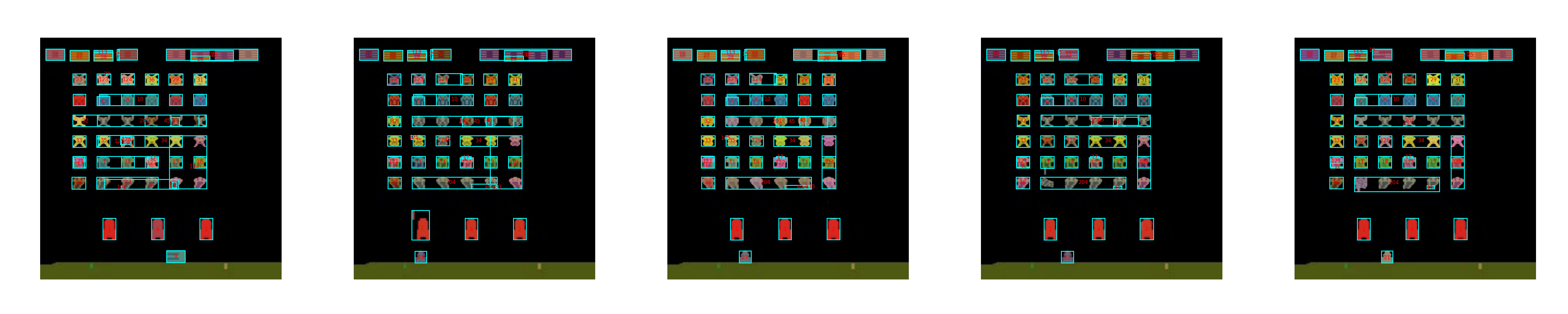}
    \caption{SpaceInvaders.}
    \label{SpaceInvaders_concatenated}
  \end{subfigure}
  \begin{subfigure}{1\linewidth}
    \includegraphics[width=\linewidth]{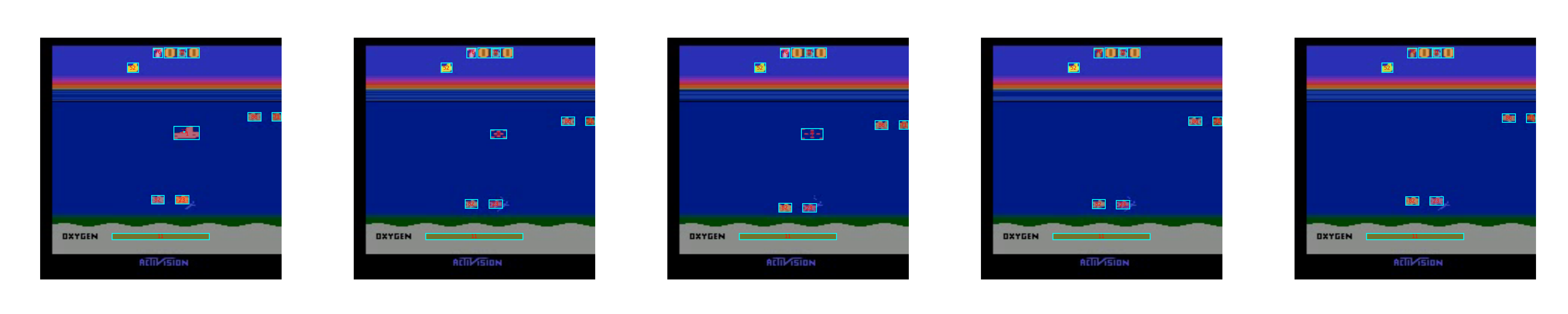}
    \caption{Seaquest.}
    \label{Seaquest_concatenated}
    \hfill
  \end{subfigure}
  \begin{subfigure}{1\linewidth}
    \includegraphics[width=\linewidth]{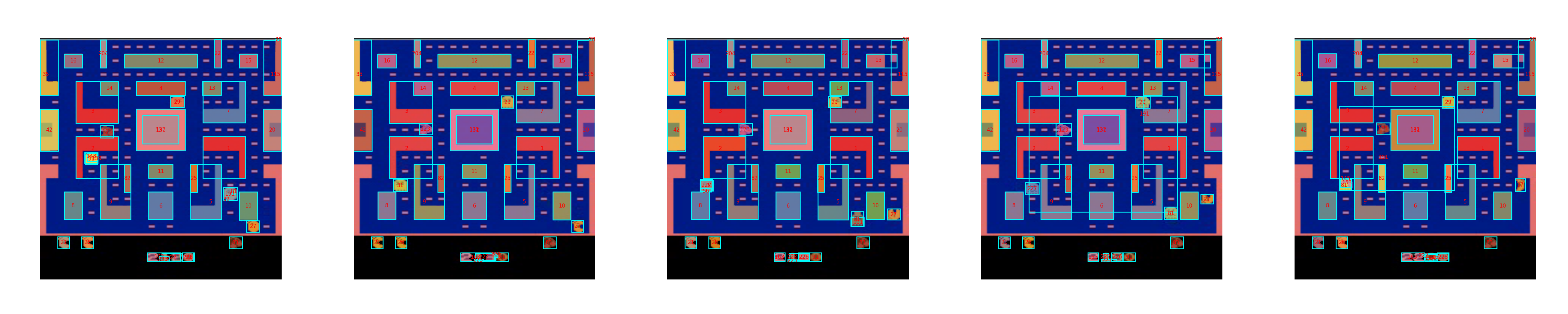}
    \caption{MsPacman.}
    \label{MsPacman_concatenated}
  \end{subfigure}
\caption{FastSAM and DeAot Extract Segmentation Results from Five Frames per Game, Sampled Every Five Frames.}
\end{figure}\label{segmentation results}

Note that our segmentation data set completes accurate segmentation of objects in most cases.

\section{Comprehensive Prompt Template Overview}\label{Comprehensive Prompt Template Overview}

In this section, the full prompt template is presented. It comprises a public description template, along with two distinct policy illustration templates. The public description template outlines the fundamental aspects of the task and the underlying policy (refer to \cref{prompt_pong_public1} and \cref{prompt_pong_public2}). The policy illustration templates are further bifurcated into policy explanation and decision explanation, detailed in \cref{prompt_pong_policy} and \cref{prompt_pong_decision}, respectively.

\clearpage 
\begin{table*}[]
\caption{The first part of pong's public prompt.}
\label{prompt_pong_public1}
\begin{tabularx}{\textwidth}{|X|} 
\hline
You need to help a user to analyze a control policy for the task Pong available in the OpenAI Gym repository. The policy is obtained with deep reinforcement learning.\\ You need to first understand the goal of the task and the policy.\\ \\ \\ \# task Description\\ \\ There are two paddles in the task screen, which are located at the left and right side of the screen. The agent controls the right paddle, and its opponent controls the left paddle. Both of them can only control the paddle to move up or down. They cannot move leftward or rightward\\ \\ Like a pingpong task, the agent competes against its opponent by stricking the ball to the opponent's side (left). The agent earns a point if its opponent fails to strick the ball back.\\ \\ The agent needs to solve the task in discrete steps. At each step, it takes as input the task screen, and it needs to take one of the three actions: \\ \\ * noop: take no operation \\ * up: move its paddle upward.\\ * down: move its paddle downward.\\ \\ \\ \# The policy\\ \\ \#\# Input Variable\\ \\ We set up a xOy-coordinate system for the task screen. The origin is at the upper left corner. The positive direction of the y-axis is downwards, and the positive direction of the x-axis is to the right. We provide the agent with the latest four consecutive frames and use the coordinates of objects in these frames as input. Frame 4 is the current frame. Frame 3 is the frame obtained at one step before. Frame 2 is the frame obtained at two steps before, and frame 1 is the frame obtained at three steps before. You can use the coordinates of the same object in different time steps to infer the motion of the object.\\ \\ The objects of interest are the agent, the opponent, and the ball. The input variables follows this naming convention: {[}x/y{]}\_object\_frame. For example, x\_agent\_1 is the x coordinate of the agent at frame 1. Remember, the input variables represent coordinates of some objects, and they are in the range {[}0,1{]}.
\\\hline
\end{tabularx}
\end{table*}

\clearpage 
\begin{table*}[]
\caption{The second part of pong's public prompt.}
\label{prompt_pong_public2}
\begin{tabularx}{\textwidth}{|X|} 
\hline
\#\# Logits\\ \\ logits\_noop1 = -0.56*y\_agent\_1**2 - 0.38*y\_agent\_1*y\_agent\_2 - 0.087*y\_agent\_1*y\_opponent\_1 - 0.16*y\_agent\_1*y\_opponent\_2 - 0.76*y\_agent\_1*y\_opponent\_3 - 0.51*y\_agent\_1*y\_opponent\_4 - 0.54*y\_agent\_1 - 0.24*y\_agent\_2**2 - 0.073*y\_agent\_2 + 0.27*y\_agent\_4**2 + 0.55*y\_agent\_4 - 0.078*y\_opponent\_1**2 - 0.33*y\_opponent\_1*y\_opponent\_2 - 0.2*y\_opponent\_1 - 0.35*y\_opponent\_2**2 - 0.5*y\_opponent\_2 - 0.34*y\_opponent\_3**2 - 0.45*y\_opponent\_3*y\_opponent\_4 - 0.32*y\_opponent\_3 - 0.15*y\_opponent\_4**2 - 0.19*y\_opponent\_4 + 1.1\\ \\ logits\_noop2 = -0.074*y\_agent\_1*y\_opponent\_2 + 0.059*y\_agent\_1*y\_opponent\_3 - 0.097*y\_agent\_4 - 0.16*y\_opponent\_1*y\_opponent\_2 - 0.18*y\_opponent\_2**2 - 0.27*y\_opponent\_2 + 0.063*y\_opponent\_4\\ \\ logits\_up1 = 0.23*y\_agent\_1**2 + 0.59*y\_agent\_1*y\_agent\_2 + 0.4*y\_agent\_2**2 + 0.11*y\_agent\_2 - 1.5*y\_agent\_4**2 - 3.6*y\_agent\_4 + 0.068*y\_opponent\_3 + 1.1\\ \\ logits\_down1 = 0.09*x\_ball\_3 + 0.12*x\_ball\_4 - 0.21*y\_agent\_1**2 + 0.12*y\_agent\_1*y\_opponent\_1 + 0.27*y\_agent\_1*y\_opponent\_2 - 0.43*y\_agent\_1*y\_opponent\_3 - 0.28*y\_agent\_1*y\_opponent\_4 + 0.13*y\_agent\_2 + 0.14*y\_agent\_4**2 + 0.43*y\_agent\_4 + 0.087*y\_ball\_3 + 0.15*y\_ball\_4 + 0.14*y\_opponent\_1**2 + 0.6*y\_opponent\_1*y\_opponent\_2 + 0.61*y\_opponent\_1 + 0.65*y\_opponent\_2**2 + 1.1*y\_opponent\_2 - 0.2*y\_opponent\_3**2 - 0.26*y\_opponent\_3*y\_opponent\_4 - 2.8*y\_opponent\_3 - 0.085*y\_opponent\_4**2 - 0.14*y\_opponent\_4 - 2.3\\ \\ logits\_up2 = 0.063*x\_ball\_4 - 0.078*y\_agent\_1 + 0.18*y\_agent\_2**2 + 0.52*y\_agent\_2*y\_agent\_3 + 0.35*y\_agent\_2*y\_opponent\_1 + 0.29*y\_agent\_2*y\_opponent\_2 + 0.26*y\_agent\_2 + 0.38*y\_agent\_3**2 + 0.51*y\_agent\_3*y\_opponent\_1 + 0.42*y\_agent\_3*y\_opponent\_2 + 1.6*y\_agent\_3 - 8.2*y\_agent\_4 - 0.085*y\_ball\_3 + 0.17*y\_opponent\_1**2 + 0.28*y\_opponent\_1*y\_opponent\_2 + 0.45*y\_opponent\_1 + 0.11*y\_opponent\_2**2 + 0.15*y\_opponent\_2 - 0.074*y\_opponent\_3 + 0.26\\ \\ logits\_down2 = -0.052*x\_ball\_1 - 0.068*x\_ball\_3 - 0.093*x\_ball\_4 + 0.18*y\_agent\_1 - 0.17*y\_agent\_2**2 - 0.49*y\_agent\_2*y\_agent\_3 - 0.33*y\_agent\_2*y\_opponent\_1 - 0.27*y\_agent\_2*y\_opponent\_2 - 0.39*y\_agent\_2 - 0.35*y\_agent\_3**2 - 0.48*y\_agent\_3*y\_opponent\_1 - 0.4*y\_agent\_3*y\_opponent\_2 - 0.38*y\_agent\_3 + 0.15*y\_agent\_4**2 + 0.54*y\_agent\_4 - 0.06*y\_ball\_1 - 0.064*y\_ball\_3 - 0.11*y\_ball\_4 - 0.17*y\_opponent\_1**2 - 0.28*y\_opponent\_1*y\_opponent\_2 - 0.58*y\_opponent\_1 - 0.13*y\_opponent\_2**2 - 0.38*y\_opponent\_2 + 2.2*y\_opponent\_3 - 0.052*y\_opponent\_4 - 3.6\\ \\ \\ \#\# The Probability of Actions\\ \\ action\_noop = {[}exp(logits\_noop1) + exp(logits\_noop2){]} / sum(exp(logits))\\ \\ action\_up = {[}exp(logits\_up1) + exp(logits\_up2){]} / sum(exp(logits))\\ \\ action\_down = {[}exp(logits\_down1) + exp(logits\_down2){]} / sum(exp(logits))
\\\hline
\end{tabularx}
\end{table*}

\clearpage 
\begin{table*}[]
\caption{Pong's policy interpretation prompt.}
\label{prompt_pong_policy}
\begin{tabularx}{\textwidth}{|X|} 
\hline
\# Your Task\\ You need to analyze this policy based on its mathematical properties. You must follow the following rules.\\ \\ 1. You can also leverage your own knowledge about the goal of the task, but the conclusions for the policies have to be based on the mathematical properties of the policy.\\ \\ 2. You need to analyze the policy in these three steps: (a) analyze how changes in variables affect action logits,  (b) analyze how changes in logits affect the probability of taking action, and (c) summarize the Influence of input variables on action probabilities.\\ \\ 3. When performing (a), remember that the input variables represent the location of an object. Take into consideration that the input variables are within {[}0,1{]}. Pay attention to the coefficients of each input variable and constants (if any).  \\ \\ 4. When performing (b), remember that the probability of actions sum to one. \\ \\ 5. An increase of the logit of certain actions might results in an increase in the probability of that action. A decrease of the logit of certain actions might results in an decrease in the probability of that action. \\ \\ 6. When performing (c), summarize your findings from (a) and (b).\\ \\ For example, for logits\_up1, first think about the coefficient of x\_ball\_2.  Since the values of  x\_ball\_2 are within {[}0,1{]}, how does it affect the logit of moving up? How does it affect the probability of moving up?\\ \\ 7. Be specific the effect of each term.\\ \\ \#\# Output\\ Organize your response as (1) equation, (2) influential variables, and (3) analysis. Render the equations into latex format. Use the object names and frame indices as subscripts. For example, y\_\textbackslash{}text\{agent,1\}. Use the name of actions as the subscript of logits. For example, logits\_\textbackslash{}text\{noop\}. Only keep two significant digits for each number.\\ \\ Now, analyze action noop.\\ \\ \{Chatgpt response\}\\ \\ Analyze action up.\\ \\ \{Chatgpt response\}\\ \\ Analyze action down.\\ \\ Provide a summary for your recent analysis. Follow the rules below.\\ \\ 1. Be specific on when will the agent chooses certain actions.\\ 2. Your summary should be consistent with your analysis.\\ 3. Organize your response in markdown format.\\ \\ Here is a recap for our set up.\\ \\ \{Recap in public prompt\} 
\\\hline
\end{tabularx}
\end{table*}

\clearpage 
\begin{table*}[]
\caption{Pong's decision explanation prompt.}
\label{prompt_pong_decision}
\begin{tabularx}{\textwidth}{|X|} 
\hline
\# Your Task\\ We used this policy to play the task and collected some data. You need to explain why the agent took a specific action when the input variables took specific values.\\ \\ The action taken by the agent is up.\\ The value of y\_ball\_1 is 0.9018810391426086\\ The gradient of the log-likelihood for action up with respect to y\_ball\_1 is 2.75e-04.\\ The value of x\_ball\_1 is 0.5703107714653015\\ The gradient of the log-likelihood for action up with respect to x\_ball\_1 is -4.13e-04.\\ The value of y\_ball\_2 is 0.5564423203468323\\ The gradient of the log-likelihood for action up with respect to y\_ball\_2 is -4.36e-04.\\ The value of x\_ball\_2 is 0.6364298462867737\\ The gradient of the log-likelihood for action up with respect to x\_ball\_2 is -7.43e-06.\\ The value of y\_ball\_3 is 0.4664875864982605\\ The gradient of the log-likelihood for action up with respect to y\_ball\_3 is -8.24e-04.\\ The value of x\_ball\_3 is 0.7508251070976257\\ The gradient of the log-likelihood for action up with respect to x\_ball\_3 is 3.87e-04.\\ The value of y\_ball\_4 is 0.7325012683868408\\ The gradient of the log-likelihood for action up with respect to y\_ball\_4 is 7.06e-04.\\ The value of x\_ball\_4 is 1.0\\ The gradient of the log-likelihood for action up with respect to x\_ball\_4 is 1.34e-03.\\ The value of y\_opponent\_1 is 1.0\\ The gradient of the log-likelihood for action up with respect to y\_opponent\_1 is 2.09e-02.\\ The value of y\_opponent\_2 is 1.0\\ The gradient of the log-likelihood for action up with respect to y\_opponent\_2 is 1.67e-02.\\ The value of y\_opponent\_3 is 1.0\\ The gradient of the log-likelihood for action up with respect to y\_opponent\_3 is -2.63e-03.\\ The value of y\_opponent\_4 is 1.0\\ The gradient of the log-likelihood for action up with respect to y\_opponent\_4 is -8.92e-04.\\ The value of y\_agent\_1 is 1.0\\ The gradient of the log-likelihood for action up with respect to y\_agent\_1 is -1.39e-02.\\ The value of y\_agent\_2 is 1.0\\ The gradient of the log-likelihood for action up with respect to y\_agent\_2 is 4.85e-03.\\ The value of y\_agent\_3 is 1.0\\ The gradient of the log-likelihood for action up with respect to y\_agent\_3 is 4.58e-02.\\ The value of y\_agent\_4 is 0.0\\ The gradient of the log-likelihood for action up with respect to y\_agent\_4 is -4.94e-02.\\ \\ \#\# Output\\ \\ You need to provide a concise explanation for why the agent took this action when the input variables took these values. \\ For example, would the agent earn a point by choosing such an action?\\ \\ There are a few rules that need to be followed.\\ 1. Your explanations should be specific. You should explain why the action up is preferred over other actions.\\ 2. Your explanations should be easy to read.\\ 3. Your explanations should be entirely based on the equations for the policy, the values of input variables, and the gradients of action log-likelihood with respect to input variables.\\ 4. Your explanations should be consistent with the definition of the input variables and the coordinate system.\\ \\ Render the equations into latex format. Use the object names and frame indices as subscripts. For example, y\_\textbackslash{}text\{agent,1\}. Use the name of actions as the subscript of logits. For example, logits\_\textbackslash{}text\{noop\}. Only keep two significant digits for each number.
\\\hline
\end{tabularx}
\end{table*}

\end{document}